\documentclass[10pt,twocolumn,letterpaper]{article}

\usepackage{subcaption}
\usepackage{iccv}
\usepackage{times}
\usepackage{epsfig}
\usepackage{graphicx}
\usepackage{amsmath}
\usepackage{amssymb}
\usepackage{multirow}
\usepackage{booktabs}
\usepackage{adjustbox}
\usepackage{comment}
\usepackage[accsupp]{axessibility}  

\newcommand{\myparagraph}[1]{\vspace{4pt}\noindent\textbf{#1}}
\usepackage[dvipsnames]{xcolor}

\newcommand\our{\text{GeoWarp}}

\usepackage[pagebackref=true,breaklinks=true,letterpaper=true,colorlinks,bookmarks=false]{hyperref}
\usepackage[all]{hypcap}  

\usepackage{my_ref}

\iccvfinalcopy 

\def\iccvPaperID{8952} 

\ificcvfinal\fi

\begin{document}

\title{Viewpoint Invariant Dense Matching for Visual Geolocalization}

\author{Gabriele Berton\textsuperscript{1,2}, Carlo Masone\textsuperscript{2}, Valerio Paolicelli\textsuperscript{2} and Barbara Caputo\textsuperscript{1,2}\\
\textsuperscript{1}Politecnico di Torino
\textsuperscript{2}Italian Institute of Technology
\\
{\tt\small [gabriele.berton, barbara.caputo]@polito.it \tt\small[carlo.masone, valerio.paolicelli]@iit.it}

}


\maketitle
\ificcvfinal\thispagestyle{empty}\fi

\begin{abstract}
In this paper we propose a novel method for image matching based on dense local features and tailored for visual geolocalization. Dense local features matching is robust against changes in illumination and occlusions, but not against viewpoint shifts which are a fundamental aspect of geolocalization. Our method, called {\our}, directly embeds invariance to viewpoint shifts in the process of extracting dense features. This is achieved via a trainable module which learns from the data an invariance that is meaningful for the task of recognizing places. 
We also devise a new self-supervised loss and two new weakly supervised losses to train this module using only unlabeled data and weak labels.
{\our} is implemented efficiently as a re-ranking method that can be easily embedded into pre-existing visual geolocalization pipelines.
Experimental validation on standard geolocalization benchmarks demonstrates that {\our} boosts the accuracy of state-of-the-art retrieval architectures.
The code and trained models are available at \url{https://github.com/gmberton/geo_warp}
\end{abstract}

\section{Introduction}
\noindent
Visual geolocalization (VG), \ie, the task of finding the position where a given photograph was taken, is a fundamental problem in numerous applications, such as robotics localization in GPS-denied environments or augmented reality.
This task is cast as an image retrieval problem where the photograph to be localized (\emph{query}) is matched against a labeled database in the space of some global image representations.
Much of the recent literature in VG has focused on improving these global representations, moving from aggregations of handcrafted local features \cite{azzi2016_lf_vpr, hays2008_lf_vpr, russel2011_lf_vpr} to more powerful and compact  CNN-based global descriptors \cite{arandjelovic2016_netvlad, kim17_crn, cao20_delg}.
Nevertheless, since global descriptors summarize the whole visual content in the image, they lack robustness to occlusions and clutter \cite{sattler2018_benchmark6DOF} and may fail to capture the similarity of two views with a small overlap \cite{kanji2016_small_overlap}.


\begin{figure}[t!]
    \centering
    \includegraphics[width=\linewidth]{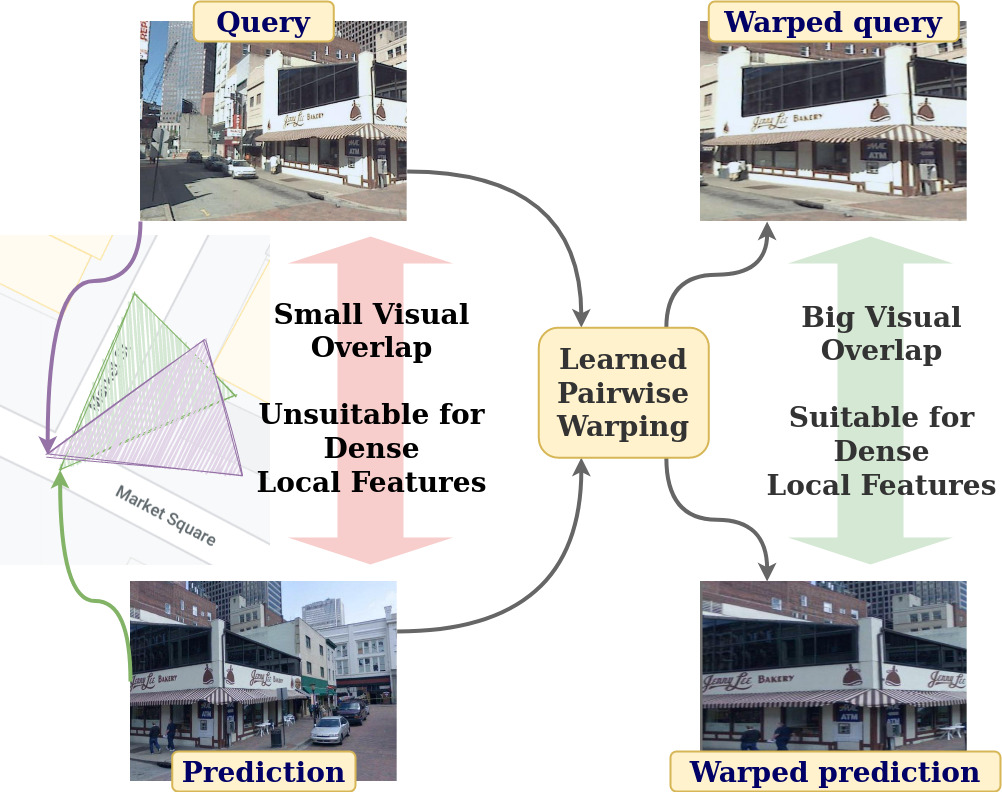}
    \vspace{2pt}
    \caption{The appearance of two different views of the same place may differ significantly, thus making it hard to match them. Our method warps both images to a closer geometrical space and then computes their similarity using deep dense local features.}
    \label{fig:teaser_p}
\end{figure}

Using sparse local invariant features to establish direct geometrical correspondences among images is an effective way to solve this problem in general visual matching tasks.
In visual geolocalization this solution is compromised by the fact that different images of the same place may have strong visual differences from one another, \eg, due to illumination or seasonal variations \cite{masone21_survey}. 
In such circumstances the detection of keypoints becomes unreliable, causing non-repeatable local invariant features \cite{torii2018_tokyo,taira2018_inloc}.
A few recent studies in visual geolocalization have demonstrated that this problem can be circumvented by removing the detection step altogether and using dense grids of local features to match places across strong visual shifts \cite{torii2018_tokyo} or with few textures \cite{taira2018_inloc}.
However, the robustness acquired by removing the detection step comes at the cost of a reduced invariance to geometric transformations.

Since viewpoint shifts are a fundamental problem of visual geolocalization, we propose a new dense matching method, called \our, that is endowed with some invariance to geometric transformations. 
Our dense matching is a trainable operation that learns an invariance that is meaningful for the task of recognizing places, in a data driven manner (see \cref{fig:teaser_p}).
Being an operation dependent on the two images to be matched, it cannot be applied database wide since it would have to be computed again for each query.
Therefore, we first perform a neighbor search on the database using state-of-the-art global descriptors, and then apply our novel dense matching on the shortlist of retrieved results to re-rank them.
On a technical level, our dense matching revolves around a new lightweight warping regression module that can be efficiently trained in a self-supervised fashion, which makes it possible to train it on unlabeled data. To further improve the results we devise two weakly supervised losses, which allow the network to gain robustness to common issues such as occlusions and appearance shifts, requiring only weak labels from each image.


\myparagraph{Contributions:}
\begin{itemize}
    \item We introduce a new dense matching method, tailored for visual geolocalization, that has an intrinsic invariance to viewpoint shifts. This dense matching can be easily integrated into standard retrieval pipelines for geolocalization.
    \item We present a new trainable method for pairwise image warping. This module is trained with three new losses: a self-supervised loss and two weakly supervised losses, that allow to rely only on unlabeled data or take advantage of weak labels, which are commonly available for visual geolocalization datasets.
    \item We present an extensive ablation and demonstrate on several standard datasets for visual geolocalization that our method significantly boosts the accuracy with a wide variety of retrieval networks.
\end{itemize}

\section{Related works}

\paragraph{Local features}
Image retrieval has been historically approached with sparse local features using hand-crafted techniques such as SIFT \cite{lowe2004_sift} and SURF \cite{bay08_surf}.
Such features are typically aggregated into fixed-length vectors such as bag-of-visual-words \cite{philbin2007_bow} or VLAD \cite{arandjelovic13_vlad} to ensure efficient similarity computation.
To overcome challenges related to variations in viewpoint, sparse local features methods often rely on spatial verification using RANSAC \cite{fischler1981_ransac}. Alternatively, \cite{torii2018_tokyo} proposes using dense local features coupled with the generation of synthetic images simulating multiple views of the same scene.
Although most CNN-based methods rely on global descriptors \cite{arandjelovic2016_netvlad, kim17_crn, ng20_solar, radenovic2019_gem, tolias2016_rmac, gordo17_qe}, recent works \cite{noh17_delf, cao20_delg} show promising results employing deep local features followed by spatial verification. 

\paragraph{Global features}
Since the advent of deep learning, image retrieval systems have been predominantly based on global features descriptors extracted by convolutional architectures. In such systems, images are passed to a convolutional encoder to extract dense local features, which are then fed into aggregation or pooling layers such as NetVLAD \cite{arandjelovic2016_netvlad} or GeM \cite{radenovic2019_gem}. Such networks are trained with ranking based losses for visual geolocalization \cite{arandjelovic2016_netvlad, kim17_crn}, whereas classification losses are commonly employed for the related task of landmark retrieval \cite{radenovic2019_gem, ng20_solar, tolias2016_rmac, gordo17_qe}.

\paragraph{Re-ranking techniques}
Re-ranking techniques are commonly used in image retrieval to re-assess the retrieved predictions produced by the retrieval system, trading computational time for a boost in accuracy.
A common method is query expansion \cite{chum2007_qe, gordo17_qe, arandjelovic2012_three_things, radenovic2019_gem}, where the results of a first search are filtered and aggregated to perform a second search.
An alternative to query expansion is given by diffusion \cite{donoser2013_diffusion, yang2019_diffusion}, a family of methods which aim at exploiting the context similarities between all elements of the database to unveil the data manifold, which is not captured by the pairwise similarity search.
Closer to our approach, other works \cite{noh17_delf, cao20_delg, taira2018_inloc, sarlin20_superglue} perform a first retrieval search using global features, following it with a post-processing step done with local features (spatial verification). Among these, the closest to our work is \cite{taira2018_inloc} which, as in our method, uses the same encoder both to generate a global descriptor for the nearest neighbor search and to extract dense local features for re-ranking. However, the dense local features are used to find an accurate camera pose through RANSAC \cite{fischler1981_ransac}, whereas we perform a learned pairwise transformation before local features extraction.

\paragraph{Geometric image transformation}
Traditionally, correspondences between pairs of images have been computed by finding points of interest and extracting local descriptors from such points \cite{bay08_surf}.
More recent works rely on features extracted from CNNs, which can be used as input for a second deep neural network \cite{brachmann2018_6d_loc, detone2016_deep_homography, rocco2018_warp, kanazawa2016_warpnet, brachmann2019_ngransac, rau2020_overlap} or to a RANSAC algorithm \cite{taira2018_inloc}. 
In particular, \cite{detone2016_deep_homography} proposes to slightly perturb the patch of an image with a homography, which is then predicted with a regression VGG-like network. 
\cite{taira2018_inloc} and \cite{brachmann2018_6d_loc} use local features followed respectively by RANSAC \cite{fischler1981_ransac} and DSAC \cite{brachmann2017_dsac} to predict the 6DOF camera pose within a given 3D environment.
\cite{kanazawa2016_warpnet} uses a Siamese CNN to predict a thin-plate spline transformation between two images of birds, while \cite{rocco2018_warp} extends the method to work on image instances from other classes than birds.
Both methods rely on a pair of images with foreground objects from the same category, with little to no occlusion, to estimate a transformation from one image to the other.
While keeping in considerations valuable lessons learned from previous works, we instead propose a pairwise transformation, aimed at morphing both input images.
This ensures clutter or unwanted elements to be removed from both of them, while being robust to pairs of images with little visual overlap.
Moreover, to ensure robustness of the network to occlusions and dynamic objects, we propose two novel weakly supervised losses, which leverage photos from the same scene taken over the years.

\section{Method} \label{subsec:Method}
\noindent
We consider the problem of geolocalizing an unseen RGB image $I_q$ given a gallery of geotagged images $\mathcal{G} = \{(I_i, z_i)\}$, where $z_i$ is the GPS coordinate of the image $I_i$.
We further assume having a training dataset of geotagged images $\mathcal{T}$, divided into training queries and training gallery.
We propose to address the visual geolocalization problem by first performing a similarity search over $\mathcal{G}$ based on global descriptors, which produces a set of predictions $\mathcal{P} \subset \mathcal{G}$. Then, we use a novel dense matching method to sort the top predictions in $\mathcal{P}$ based on a similarity measure with the query computed from dense local descriptors.

\subsection{Place retrieval with global descriptors} \label{sec:global_descriptors}
\noindent
As a first step, our method implements a classic pipeline for place retrieval based on global image descriptors.
To generate the global descriptors we utilize a CNN that is composed of two elements:
\begin{itemize}
    \item A convolutional encoder $E$, that takes an image and outputs a tensor $f \in \mathbb{R}^{h_f \times w_f \times C}$. The tensor $f$ can be interpreted as a dense $h_f \times w_f$ grid of $C$-dimensional local feature descriptors and we denote the local feature at the spatial position $(i, j)$ of the grid as $\boldsymbol{f}(i,j)$;
    \item A layer $A$ that takes the tensor $f$ and produces a vectorial representation of the image either by aggregation (\eg, NetVLAD \cite{arandjelovic2016_netvlad}) or by pooling (\eg, GeM \cite{radenovic2019_gem}).
\end{itemize}
Namely, given an image $I$, its global descriptor is $A(E(I))$. This network is trained specifically for place retrieval using a triplet loss and following the protocol from \cite{arandjelovic2016_netvlad}.
At inference time, given a new query $I_q$ we perform a nearest neighbour search over the gallery $\mathcal{G}$ in the space of the global descriptors, which yields the set of predictions $\mathcal{P}$.

\begin{figure*}[t!]
    \centering
    \includegraphics[width=\linewidth]{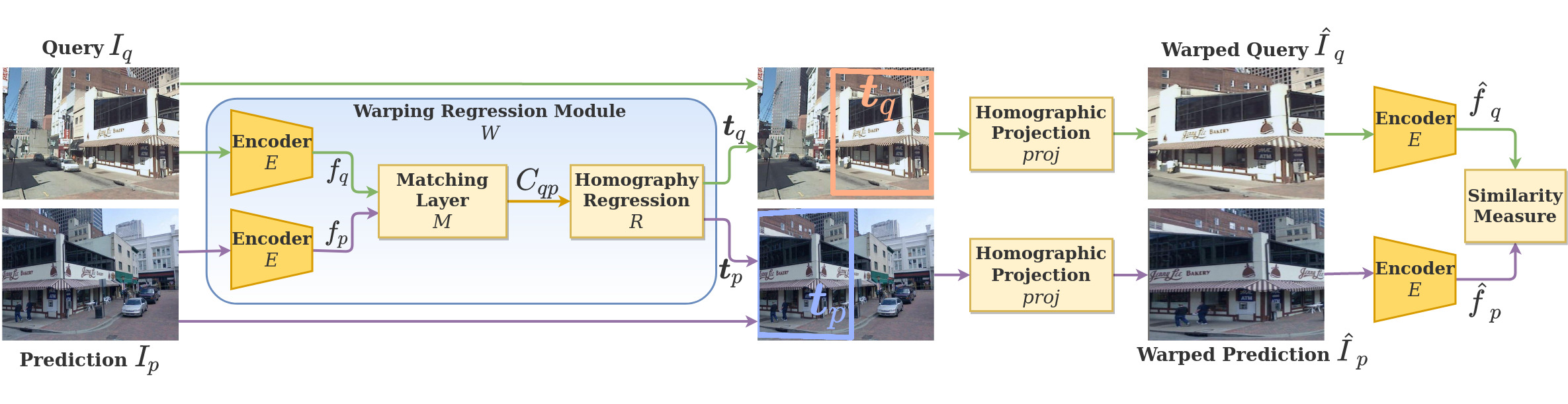}
    \caption{Diagram of our architecture's functioning at inference time.
    The warping regression module is entitled with estimating two quadrilaterals $\boldsymbol{t}_q$ and $\boldsymbol{t}_p$ from the two images on the left (query and prediction). The images are then warped with a homography, and their similarity is computed on their deep dense local features.
    Note that an ideally perfect warping which generates very similar images would be counterproductive, as images taken far away (i.e. a query-negative pair) would end up with a similar features representation.
    }
    \label{fig:warp_inference}
\end{figure*}

\subsection{Re-ranking with dense local descriptors}
\noindent
We propose to re-rank the predictions $\mathcal{P}$ by recomputing their similarity to the query $I_q$ using dense local feature descriptors. Although sparse invariant local features are successfully used in various visual matching problem, in visual geolocalization they have shown limited reliability due to the challenging visual conditions that can cause failures in the keypoint detection \cite{zhang2020_reference_pg}. On the other hand, directly matching densely sampled local features has shown great promise \cite{torii2018_tokyo,taira2018_inloc} at the cost of a limited invariance to viewpoint shifts. 
To overcome this limitation we re-rank the predictions $\mathcal{P}$ using a new trainable matching operation, tailored for visual geolocalization, that learns to extract dense local features that, to some extent, are invariant to viewpoint shifts (see \cref{fig:warp_inference}).

To endow the feature extraction process with invariance to viewpoint shifts, we propose a warping regression module $W$ that takes the query $I_q$ and a prediction $I_p \in \mathcal{P}$ and estimates a homographic transformation for each of the two images. The mapping from images to homographic transformations is learned from the data  with the goal to better align two different views of a same scene, even when they  have limited overlap. 
The regression module will be detailed in \cref{subsec:Pairwise image warping}. For now, we indicate its mapping as
\begin{equation}
    \label{eq:warping_generic}
    W(I_q, I_p) = [\boldsymbol{t}_q, \boldsymbol{t_p}]
\end{equation}
where $\boldsymbol{t}_q \in \mathbb{R}^8$ and $\boldsymbol{t}_p \in \mathbb{R}^8$ are the estimated parameters for the transformations: they can be seen as the four points needed to extract a homography matrix for the eight-point transformation \cite{hartley97_eight_points}, with the remaining four points being the corners of the image (see \cref{fig:warp_inference}).
With the estimated parameters $\boldsymbol{t}_q$ and $\boldsymbol{t}_p$, the images $I_q$ and $I_p$ can be transformed using the well known eight-points transformation \cite{hartley97_eight_points} to generate two warped images $\hat{I}_q$ and $\hat{I}_p$. Hereinafter we will denote this transformation as
\begin{equation}
    \label{eq:proj}
    \begin{aligned}
        \hat{I}_q &= \text{proj}(I_q, \boldsymbol{t}_q)\\
        \hat{I}_p &= \text{proj}(I_p, \boldsymbol{t}_p)
    \end{aligned}
\end{equation}
As it will be discussed in \cref{subsec:warp_training}, this learnable transformation is trained to bring two overlapping views of the same location to a closer perspective. However, when the two images depict a different scene, the limited effect of the projection does not affect the final proximity of both images. Qualitative results of the projection are shown in Fig. \ref{fig:qualitative_results} and in the supplementary material.

Finally, we extract dense local features from the warped images $\hat{I}_q$ and $\hat{I}_p$, by means of the same encoder $E$ that was trained for producing global image representations (\cref{sec:global_descriptors}), i.e., 
\begin{equation}
    \begin{aligned}
        \hat{f}_q &= E(\hat{I}_q)\\
        \hat{f}_p &= E(\hat{I}_p)
    \end{aligned}
\end{equation}
Using the same encoder $E$ that was trained specifically for place retrieval helps to produce features that are highly discriminative for locations.
The output of the proposed operation is a similarity score between the query and prediction using the distance between the local features of their warped counterparts, i.e.,
\begin{align}
    \label{eq:rerank_score}
    d_{p} = \sum_{i=0}^{w_f-1} \sum_{j=0}^{h_f-1} \hat{\boldsymbol{f}}_q (i,j)^T \, \hat{\boldsymbol{f}}_p (i,j)
\end{align}

This procedure is repeated for all the predictions in $\mathcal{P}$, which are finally sorted according to the scores yielded by \cref{eq:rerank_score}.
This makes the time complexity of our re-ranking method $O(|\mathcal{P}|)$. Since the number of predictions that needs to be processed is usually small ($|\mathcal{P}| \ll |\mathcal{G}|$), the approach is suitable for large-scale geolocalization problems.

\subsection{Warping regression module} \label{subsec:Pairwise image warping}

\noindent
The warping regression module $W$ in \cref{eq:warping_generic} was inspired by \cite{rocco2018_warp}, albeit we limit it to homographies. Homographies aim at describing the relationship between representations of co-planar points in a scene as projected onto different viewing planes. In visual geolocalization planar surfaces are abundant, as buildings' facades often represent the most discriminative characteristic of a place, making homography ideal for our purpose.
$W$ consists of three steps (Fig. \ref{fig:warp_inference}).
First, we extract the features from the two images using the encoder $E$, \ie,
\begin{equation}
    \label{eq:features_qp}
    \begin{aligned}
        f_q &= E(I_q)\\
        f_p &= E(I_p)
    \end{aligned}
\end{equation}
Note that $E$ has a threefold purpose: it is used in the global descriptors extraction, the warping module and the final similarity score computation (\ref{eq:rerank_score}). 

Then, the features are fed to a matching layer $M$ that computes the correlation map $c_{qp} \in \mathbb{R}^{h_f \times w_f \times (h_f \times w_f)}$ between each pair of local feature descriptors from $f_q$ and $f_p$, \ie,
\begin{equation}
    c_{qp}(i,j,k) = \boldsymbol{f}_q(i,j)^T \boldsymbol{f}_p(i_k, j_k)
\end{equation}
For the sake of brevity, we write the matching operation as
\begin{equation}
    \label{eq:matching}
    M(f_q, f_p) = c_{qp}
\end{equation}
We remark that this layer is differentiable and parameterless and we refer to \cite{rocco2018_warp} for further details.

Finally, the correlation map $c_{qp}$ is given as input to a convolutional network $R$ that is designed to estimate the transformations of the two images.
In particular, the network $R$ estimates 4 points on each of the two images, \ie, a 16D vector. Formally, we denote this operation as
\begin{equation}
    \label{eq:regression}
    R(c_{qp}) = [\underbrace{\boldsymbol{p}_{q1}, \ldots, \boldsymbol{p}_{q4}}_{\boldsymbol{t}_q}, \underbrace{\boldsymbol{p}_{p1}, \ldots, \boldsymbol{p}_{p4}}_{\boldsymbol{t}_p}] = [\boldsymbol{t}_q, \boldsymbol{t}_p] \in \mathbb{R}^{16}
\end{equation}
where $\boldsymbol{p}_{q1}, \ldots, \boldsymbol{p}_{q4}$ are four points on $I_q$, $\boldsymbol{p}_{p1}, \ldots, \boldsymbol{p}_{p4}$ are four points on $I_p$, and the notation $[\boldsymbol{t}_q, \boldsymbol{t}_p]$ denotes the concatenation of the two vectors (see Fig. \ref{fig:warp_inference}).

Combining \cref{eq:features_qp,eq:matching,eq:regression} the warping regression \cref{eq:warping_generic} is summarized as,
\begin{equation}
    \label{eq:warping_detailed}
    W(I_q, I_p) = R\big( M(E(I_q), E(I_p)) \big) = [\boldsymbol{t}_q, \boldsymbol{t}_p]
\end{equation}

Although our warping regression module $W$ is inspired by \cite{rocco2018_warp}, it introduces some notable novelties. The first difference arises from the use case of the warping operation. While \cite{rocco2018_warp} regresses a geometric transformation for generic image matching, we focus specifically on the geolocalization problem.  As already mentioned, we use the same encoder $E$ that was trained for place retrieval, which means that we do not need to train a second encoder and that the extracted features encode more discriminative information for distinguishing places.

The second and most important difference is that the transformation module in \cite{rocco2018_warp} is designed to estimate only the transformation of one image while keeping the other one unchanged. On the contrary, our solution considers the more general problem in which both images can be transformed. In the spirit of deep learning, we let the module itself learn from the data whether and how much each of the two images should be transformed. This ensures greater flexibility and the possibility of achieving greater similarity between the generated pair. 
This is demonstrated quantitatively in the experiments presented in \cref{sec:experiments}, and qualitatively in Fig. \ref{fig:qualitative_results} and in the supplementary material.
This second difference also implies that the training procedure from \cite{rocco2018_warp} is not applicable to our case. Therefore, we propose a new training protocol, which is discussed next in \cref{subsec:warp_training}.


\subsection{Training the warping regression module} \label{subsec:warp_training}
\noindent
To train the warping regression module $W$ in a fully supervised way we would need a dataset with training quadruplets $\{I_a, I_b, \boldsymbol{t}_a, \boldsymbol{t}_b\}$, where $I_a$ and $I_b$ are two images of the same location viewed from different viewpoints and $\boldsymbol{t}_a$ and $\boldsymbol{t}_b$ are the ground truth parameters of the homographic transformations.
Given the lack of such a dataset, we propose a training procedure that combines a new self-supervised loss $L_{ss}$ (\cref{subsec:Self-supervised training}), as well as two novel weakly supervised losses $L_{fw}$ and $L_{cons}$ (\cref{subsec:Weakly supervised losses}).
Hence, the total loss is
\begin{equation}
    \label{eq:total_loss}
    L_{total} = \lambda_{ss} L_{ss} + \lambda_{fw} L_{fw} + \lambda_{cons} L_{cons}
\end{equation}

Before proceeding with detailing the terms in \cref{eq:total_loss}, we point out that: i) the matching layer $M$ is parameterless, and ii) the encoder $E$ was previously trained for place retrieval and its parameters are kept frozen when training $W$.
This second point also ensures that when training $W$ we can rely on features that are optimized for the geolocalization task.

\begin{figure}
    \centering
    \begin{minipage}{.11\textwidth}
        \begin{subfigure}{\textwidth}
            \includegraphics[width=\textwidth]{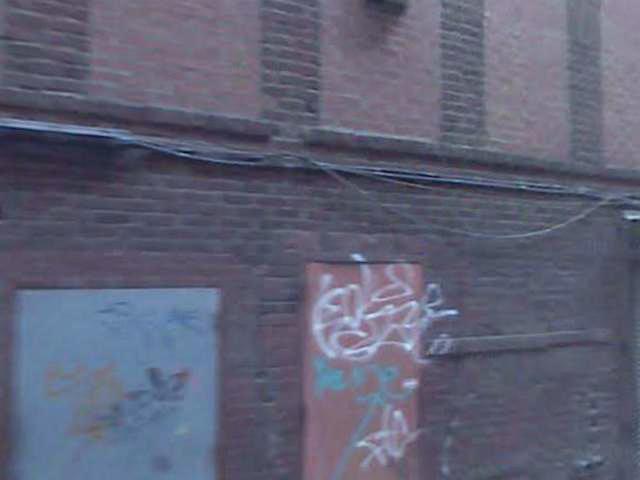}
        \end{subfigure}
        \begin{subfigure}{\textwidth}
            \includegraphics[width=\textwidth]{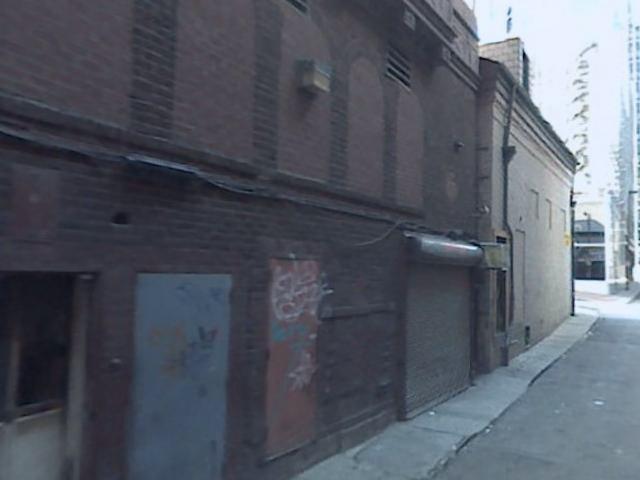} 
        \end{subfigure}
    \end{minipage}
    \begin{minipage}{.11\textwidth}
        \begin{subfigure}{\textwidth}
            \includegraphics[width=\textwidth]{imgs/cmp_base1130_3_p.jpg}
        \end{subfigure}
        \begin{subfigure}{\textwidth}
            \includegraphics[width=\textwidth]{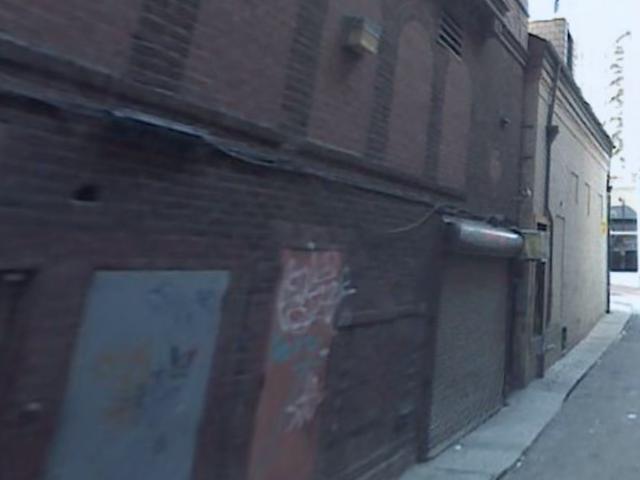} 
        \end{subfigure}
    \end{minipage}
    \begin{minipage}{.11\textwidth}
        \begin{subfigure}{\textwidth}
            \includegraphics[width=\textwidth]{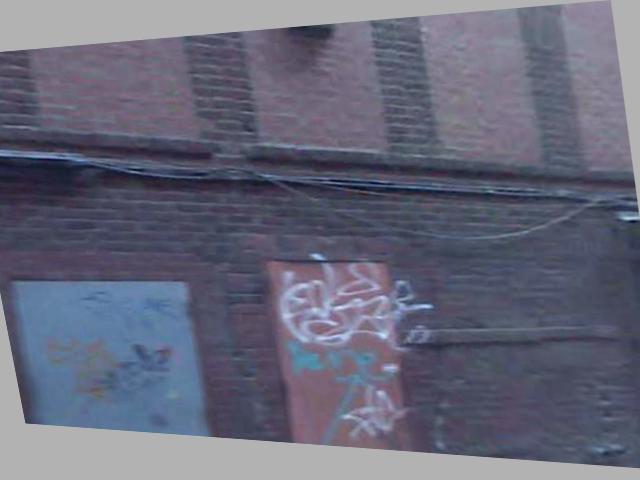}
        \end{subfigure}
        \begin{subfigure}{\textwidth}
            \includegraphics[width=\textwidth]{imgs/cmp_base1130_3_q.jpg}
        \end{subfigure}
    \end{minipage}
    \begin{minipage}{.11\textwidth}
        \begin{subfigure}{\textwidth}
            \includegraphics[width=\textwidth]{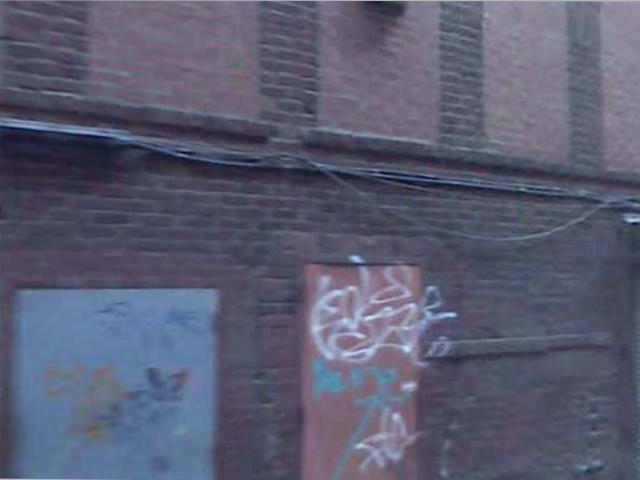}
        \end{subfigure}
        \begin{subfigure}{\textwidth}
            \includegraphics[width=\textwidth]{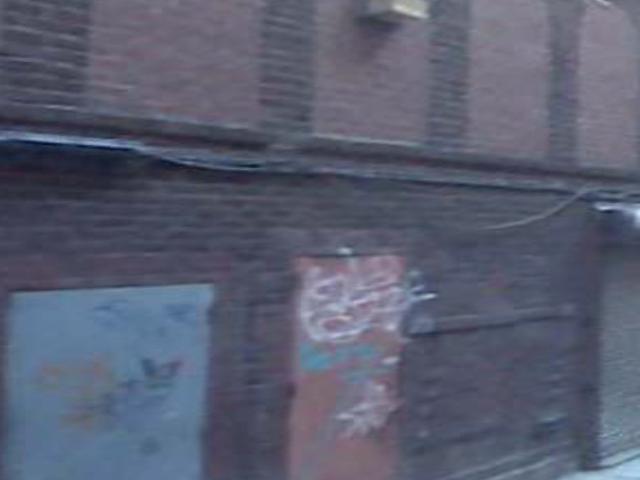}
        \end{subfigure}
    \end{minipage}
    \vspace{5pt}
    \caption{Qualitative results: the first column represents a query-prediction pair, the second column shows warping on the prediction using \cite{rocco2018_warp}, the third shows warping on the query using \cite{rocco2018_warp}, and the rightmost column is our pairwise warping.}
    \label{fig:qualitative_results}
\end{figure}


\subsubsection{Self-supervised training} \label{subsec:Self-supervised training}
\noindent
We propose to generate training quadruplets $\{I_a, I_b, \boldsymbol{t}_a, \boldsymbol{t}_b\}$ from a single training image and train the regression network in $W$ in a self-supervised fashion.
Let us consider a generic training image $I$ with shape $w \times h$. We define a procedure to randomly sample the corners of a quadrilateral on $I$:

\begin{equation}
    \label{eq:points_sampling}
    \begin{aligned}
        \boldsymbol{p}_{1} &= \left[\tfrac{U(0,k)*w}{2}, \tfrac{U(0,k)*h}{2}\right]\\
        \boldsymbol{p}_{2} &= \left[w-\tfrac{U(0,k)*w}{2}, \tfrac{U(0,k)*h}{2}\right]\\
        \boldsymbol{p}_{3} &= \left[w-\tfrac{U(0,k)*w}{2}, h-\tfrac{U(0,k)*h}{2}\right]\\
        \boldsymbol{p}_{4} &= \left[\tfrac{U(0,k)*w}{2}, h-\tfrac{U(0,k)*h}{2}\right]\\
        \text{s.t.}& \quad
        \begin{matrix}
            \boldsymbol{p}_{1}[0] = \boldsymbol{p}_4[0]\\
            \boldsymbol{p}_{2}[0] = \boldsymbol{p}_3[0]
        \end{matrix}
    \end{aligned}
\end{equation}
where $U(a, b)$ is the uniform distribution and $k \in [0, 1]$ is a constant. 
When $k$ approaches 0 the four points are close to the corners of $I$, whereas higher values move them towards the center of the image. An example with $k = 0.8$ is shown in \cref{fig:warp_ss_training}, and more examples with other values of $k$ are shown in the supplementary material.
The two constraints in \cref{eq:points_sampling} impose that the four points delimit a quadrilateral with two vertical parallel sides.
This introduces a bias towards vertically aligned images, which represent the standard in visual geolocalization datasets \cite{arandjelovic13_vlad, torii2018_tokyo, warburg2020_msls, chen2011_san_francisco, mirowski2018_streetlearn, zamir2014_gsv, berton2021_WACV, robotcar}, as well as in real world application (e.g. autonomous vehicles imagery and user-generated photos).
\begin{figure}[t!]
    \centering
    \includegraphics[width=0.8\columnwidth]{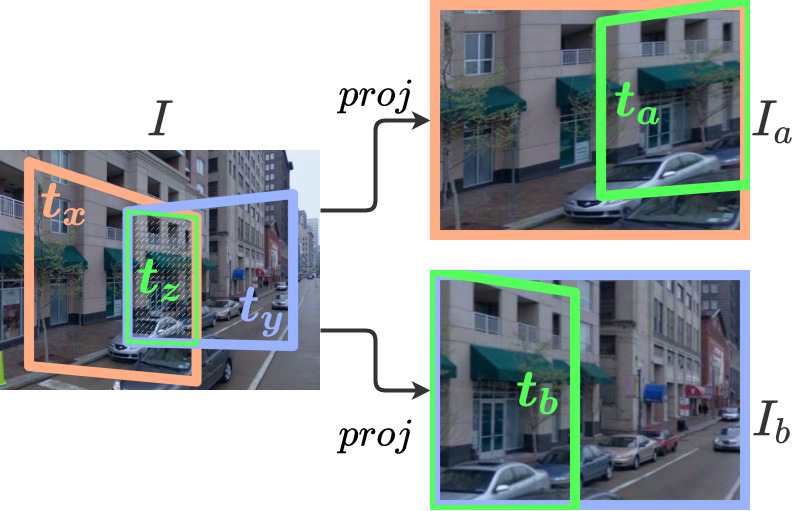}
    \caption{Self-supervised generation of training quadruplets $\{I_a, I_b, \boldsymbol{t}_a, \boldsymbol{t}_b\}$ from a single training image $I$ (left). By construction $\boldsymbol{t}_z$ is known, and so are its projections $\boldsymbol{t}_a$ and $\boldsymbol{t}_b$.}
    \label{fig:warp_ss_training}
\end{figure}
We apply the procedure \cref{eq:points_sampling} twice to generate two trapezoids on $I$, which we denote as $\boldsymbol{t}_x = [\boldsymbol{p}_{x1}, \, \boldsymbol{p}_{x2}, \, \boldsymbol{p}_{x3}, \, \boldsymbol{p}_{x4}] \in \mathbb{R}^8$ and $\boldsymbol{t}_y = [\boldsymbol{p}_{y1}, \, \boldsymbol{p}_{y2}, \, \boldsymbol{p}_{y3}, \, \boldsymbol{p}_{y4}] \in \mathbb{R}^8$.
By construction, the intersection $\boldsymbol{t}_x \cap \boldsymbol{t}_y$, is never empty and we define $\boldsymbol{t}_z = [\boldsymbol{p}_{z1}, \boldsymbol{p}_{z2}, \boldsymbol{p}_{z3}, \boldsymbol{p}_{z4}]$ as the widest trapezoid with two vertical edges within the intersection (see Fig. \ref{fig:warp_ss_training}).

Then, we generate the training quadruplet from $I$, $\boldsymbol{t}_x$,  $\boldsymbol{t}_y$ and $\boldsymbol{t}_z$ by homographic projection:
\begin{equation}
    \label{eq:ss_proj}
    \begin{aligned}
        I_a &= \text{proj}(I, \boldsymbol{t}_x) \quad &\boldsymbol{t}_a = \text{proj}(\boldsymbol{t}_z, \boldsymbol{t}_x)\\
        I_b &= \text{proj}(I, \boldsymbol{t}_y) \quad &\boldsymbol{t}_b = \text{proj}(\boldsymbol{t}_z, \boldsymbol{t}_y)
    \end{aligned}
\end{equation}
Note that in \cref{eq:ss_proj} we have used the same notation that was introduced in \cref{eq:proj} to indicate both the homographic projection of an image and the projection of a set of four points. Albeit not accurate, this prevents further encumbering the notation.

Finally, we define the self-supervised warping loss as
\begin{equation}
    \label{eq:loss_ss}
    L_{SS} = \lVert W(I_a, I_b) - [\boldsymbol{t}_a, \boldsymbol{t}_b] \rVert ^2
\end{equation}
This loss guides the network to learn to estimate the points describing the area where the two input images intersect.


\subsubsection{Weakly supervised losses}\label{subsec:Weakly supervised losses}

\begin{figure}
    \centering
    \includegraphics[width=\linewidth]{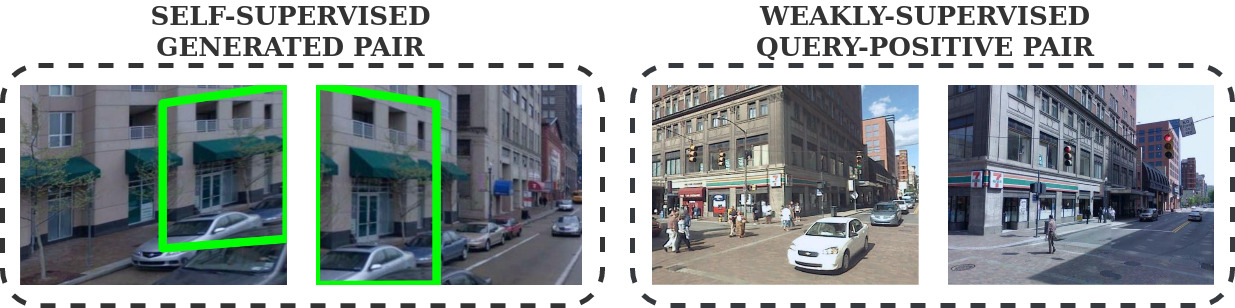}
    \caption{Examples of data used for training. With self-supervised generated images we have the advantage of knowing the intersection's ground truth, while the use of weakly supervised query-positive pairs better simulates a test-time situation.}
    \label{fig:ss_vs_ws_images}
\end{figure}

\noindent
The synthetic quadruplets produced by the self-supervised method presented in \cref{subsec:Self-supervised training} can be used effectively to train the network, however they do not provide a realistic representation of the data distribution presented at inference time by query-predictions pairs.
The synthetic images, being extracted from a single image, contain the same dynamic objects (such as vehicles and pedestrians) as well the same textures (\ie, same color of the sky, same vegetation). On the contrary, the pairs of queries and predictions seen at inference time are photos taken at different times, sometimes months or years apart.
This might result in an accuracy drop during inference.
To mitigate this unwanted behaviour, we propose to use pairs of queries and predictions from the training set $\mathcal{T}$, which we mine in a weakly supervised fashion (see \cref{fig:ss_vs_ws_images}).
Formally, we form these pairs by taking all pairs of training queries $\{(I_q, z_q)\}$ and training gallery samples $\{(I_g, z_g)\}$ that satisfy these constraints:
\begin{equation}
    \label{eq:constraint1}
    d_{geo}(z_q, z_g) < t_{geo}
\end{equation}
\begin{equation}
    \label{eq:constraint2}
    \lVert A(E(I_q)), A(E(I_g)) \rVert ^2 < t_{feat}
\end{equation}
where $d_{geo}(a, b)$ is the geographical distance between the location of two images, \ie, the distance in meters computed from their tagged GPS coordinates, $t_{geo}$ is a distance threshold (usually set to 25 meters) and $t_{feat}$ is a threshold in the features space.
Constraints \cref{eq:constraint1,eq:constraint2} impose that the image from the gallery is taken in the proximity of the query, and that the two are similar in the features space, making it very unlikely for the two images to contain a different scene.
This mining is performed once, before the start of the training, and it generates a set of pairs of images $\{(I_{q}, I_{g})\}$, where $I_{q}$ is a training query and $I_{g}$ is its related positive.
Note that the same query image might be represented more than once in the set, while others might not belong to it.
We use this set of training pairs in conjunction with two new weakly supervised losses.

\paragraph{Features-wise loss}
The first weakly supervised loss is designed to ensure that the features extracted from the training pair after warping are as close as possible, as this is the goal of the matching operation.
The features-wise loss is constructed as follows. First, we regress the homographies from the pair of images using the warping regression module $W$, warp them and extract their features, \ie,
\begin{equation}
    \begin{aligned}
        W(I_q, I_g) &= [\boldsymbol{t}_q, \boldsymbol{t}_g]\\
        \hat{f}_{q} &= E\left(\text{proj}(I_q, \boldsymbol{t}_q)\right) \\
        \hat{f}_{g} &= E\left(\text{proj}(I_g, \boldsymbol{t}_g)\right)
    \end{aligned}
\end{equation}
Afterwards, we compute the features-wise loss as
\begin{equation}
    L_{fw} = \sum_{i=0}^{w_f-1} \sum_{j=0}^{h_f-1} \left(\hat{\boldsymbol{f}}_{q} (i,j)^T \, \hat{\boldsymbol{f}}_{g} (i,j)\right)^2
\end{equation}
While the self-supervised loss \cref{eq:loss_ss} might rely on dynamic objects or lighting conditions to learn an estimate of the warping, this loss guarantees the output to be robust to such scene variations.

The loss $L_{fw}$ has some similarities with the VGG-based perceptual loss \cite{gatys2016_style}, as both losses aim at reducing the distance between two images after projecting them into features spaces. However, differently from the perceptual loss, our $L_{fw}$ computes a similarity between the two images using only high level features that are extracted from the last convolutional layer of the encoder. Since the encoder is previously trained on a visual geolocalization task, this ensures that the features are mostly extracted from parts of the image which are informative for this task. 
In this way the network's generated homography will focus mostly on static objects, whereas using a perceptual loss would force the network to learn a homography based on the entire scene represented in the image.


\paragraph{Consistency loss}
Given the lack of homography labels in our task, we propose to use self-generated pseudo-labels as ground truths to further improve the robustness.
We generate the pseudo-labels as follows:
\begin{equation}
    \label{eq:cons_los_cases}
    \begin{aligned}
        \frac{1}{N}\sum_{i=1}^N \tau_i^{-1}(W(\tau_i(I_q), \tau_i(I_g))) &= [\boldsymbol{t}_q', \boldsymbol{t}_g'] \\
        \frac{1}{N}\sum_{i=1}^N \tau_i^{-1}(W(\tau_i(I_g), \tau_i(I_q))) &= [\boldsymbol{t}_g'', \boldsymbol{t}_q''] \\
        [\boldsymbol{t}_q^*, \boldsymbol{t}_g^*] = \frac{[\boldsymbol{t}_q', \boldsymbol{t}_g'] + [\boldsymbol{t}_q'', \boldsymbol{t}_g'']}{2}
    \end{aligned}
\end{equation}
where $\tau$ is used to indicate an invertible geometric transformation applied to the images, $\tau^{-1}$ is its inverse applied to the estimated points, $N$ represents the number of transformations to apply, and $[\boldsymbol{t}_q^*, \boldsymbol{t}_g^*]$ are the generated pseudo-labels for the required homography.
Note how using a higher number of transformations ($N$) guarantees for the pseudo-labels to be more accurate (\ie closer to the ground truths), at the expense of computational requirements.
After detaching $[\boldsymbol{t}_q^*, \boldsymbol{t}_g^*]$ from the computational graph, we define the consistency loss as follows:
\begin{equation}
\begin{split}
    L_{cons} = \frac{1}{2N} \sum_{i=1}^N
    \left(\tau_i^{-1}(W(\tau_i(I_q), \tau_i(I_g))) - [\boldsymbol{t}_q^*, \boldsymbol{t}_g^*] \right) ^2
    + \\
    \left(\tau_i^{-1}(W(\tau_i(I_g), \tau_i(I_q))) - [\boldsymbol{t}_g^*, \boldsymbol{t}_q^*] \right) ^2
\end{split}
\end{equation}
A trivial solution for minimizing the consistency loss $L_{cons}$ is to teach the warping regression module to extract points at the corner of the images, effectively making the whole warping regression module and homographic projection an identity transformation. This is clearly seen in \cref{tab:ablation_losses}, where it is shown that $L_{cons}$ on its own brings no advantage, while the improvements are noticeable when it is combined with the other losses.

\section{Experiments}
\label{sec:experiments}

\subsection{Setup}

\myparagraph{Datasets and metric}
We evaluate our method on two standard visual geolocalization datasets: Pitts30k \cite{arandjelovic2016_netvlad} and R-Tokyo \cite{warburg2020_msls}.
%
The metric used to validate the results is the recall@1 as defined in \cite{arandjelovic2016_netvlad}, \ie, the percentage of queries for which the first prediction is not further than a certain threshold. In our experiments we show results for commonly used thresholds \cite{arandjelovic2016_netvlad, kim17_crn, sattler2018_benchmark6DOF, torii2015_repetitive_structures, arandjelovi2014_disLocation}, namely 10m, 25m, and 50m, corresponding to different accuracy requirements (from finer to coarser).


\myparagraph{Implementation details}
We test our dense matching on top of various retrieval systems, using different encoders (AlexNet \cite{krizhevsky2012_alexnet}, VGG16 \cite{simonyan2015_vgg} and ResNet50 \cite{he2016_resnet}) and aggregation layers (GeM \cite{radenovic2019_gem} and NetVLAD \cite{arandjelovic2016_netvlad}).
%
%
Prior to the matching layer, the feature maps produced by the encoder are resized to $15 \times 15 \times C$, where $C$ is the number of channels at the last convolutional layer, to ensure that the method works with images of different sizes.
The correlation map outputted by the matching layer is then passed to the homography regression, implemented through a sequence of six convolutional layers and a fully connected layer with output dimensionality 16. 
This final layer is initialized with weights set to 0, and with biases such that the initial estimated points (prior to any training) correspond to the four corners of the input images.
We use a batch size of 16 generated image pairs for the $L_{ss}$ and 16 query-positive pairs for the weakly supervised losses.
We train the warping regression module for 50000 iterations.
Regarding the consistency loss $L_{cons}$, we use $N=2$ geometric transformations, namely the identity transformation and an horizontal flip.
We set $t_{geo}$ to 25 meters, $t_{feat}=1.2$, $|\mathcal{P}|=5$, and the warping coefficient $k$ to 0.6 (see an ablation on $k$ in Fig. \ref{fig:ablation_k}).
Finally, we set $\lambda_{ss}=1$, $\lambda_{fw}=10$, and $\lambda_{cons}=0.1$.

\subsection{Results}
\noindent
The results of the experiments are reported in \cref{tab:results_baselines}.
We see that {\our} achieves a substantial improvement with all the configurations of encoder and aggregation methods.
Notably, the impact of {\our} is stronger with GeM than with NetVLAD.
In general, we see that NetVLAD provides better embeddings than GeM, thus reducing the impact of our method. However, this comes at the cost of using global descriptors that are 64 times heavier than GeM's, making it unsuitable for large-scale problems. Considering that {\our}'s computation time does not depend on the size of the gallery, the results achieved with a compact descriptor as GeM proves that our solution is viable for large-scale problems.

\begin{table}[t!]
\begin{adjustbox}{width=\linewidth}
\centering
\begin{tabular}{clcccccccc}
\toprule
\multirow{2}{*}{Backbone} & \multicolumn{1}{c}{\multirow{2}{*}{Method}} & \multicolumn{3}{c}{Pitts30k} & & \multicolumn{3}{c}{R-Tokyo} \\ \cline{3-5} \cline{7-9}
& \multicolumn{1}{c}{}                        & 10m  & 25m  & 50m  & & 10m  & 25m  & 50m  \\ \hline
AlexNet & GeM                                 & 50.4 & 65.3 & 70.7 & & 19.2 & 23.2 & 24.2 \\
AlexNet & GeM + GeoWarp \textbf{(ours)}       & \textbf{61.5} & \textbf{74.7} & \textbf{78.4} & & \textbf{27.9} & \textbf{34.0} & \textbf{34.7} \\ \hline
AlexNet & NV                             & 64.8 & 78.7 & 83.1 & & 41.8 & 45.8 & 46.8 \\ 
AlexNet & NV + GeoWarp \textbf{(ours)}   & \textbf{67.6} & \textbf{80.8} & \textbf{84.4} & & \textbf{44.1} & \textbf{48.8} & \textbf{49.5} \\ \hline \hline
VGG16 & GeM                                   & 55.4 & 70.6 & 76.3 & & 33.7 & 40.4 & 45.8 \\ 
VGG16 & GeM + GeoWarp \textbf{(ours)}         & \textbf{65.3} & \textbf{79.2} & \textbf{83.1} & & \textbf{49.2} & \textbf{55.9} & \textbf{58.6} \\ \hline
VGG16 & NV                               & 67.2 & 82.5 & 86.5 & & 50.2 & 56.9 & 59.9 \\
VGG16 & NV + GeoWarp \textbf{(ours)}     & \textbf{70.2} & \textbf{83.3} & \textbf{86.7} & & \textbf{54.5} & \textbf{61.6} & \textbf{63.6} \\ \hline \hline
ResNet-50 & GeM                               & 68.3 & 81.4 & 84.4 & & 36.0 & 41.8 & 45.5 \\ 
ResNet-50 & GeM + GeoWarp \textbf{(ours)}     & \textbf{70.4} & \textbf{82.7} & \textbf{85.6} & & \textbf{44.1} & \textbf{49.5} & \textbf{51.9} \\ \hline
ResNet-50 & NV                           & 70.2 & 84.3 & 87.3 & & 65.0 & 72.4 & 74.4 \\
ResNet-50 & NV + GeoWarp \textbf{(ours)} & \textbf{72.1} & \textbf{84.8} & \textbf{87.8} & & \textbf{69.7} & \textbf{74.4} & \textbf{75.4} \\ \bottomrule
\end{tabular}
\end{adjustbox}
\vspace{2pt}
\caption{Recall@1 of well-established baseline methods with and without \our. NV stands for NetVLAD \cite{arandjelovic2016_netvlad}.}
\label{tab:results_baselines}
\end{table}

\subsection{Comparison with other methods}
\noindent
We compare the performance of {\our} to state-of-the-art re-ranking solutions for visual geolocalization and landmark recognition. 
%
%
First, we consider two popular techniques for image retrieval problems, query expansion (QE) and diffusion. For QE, we use the official implementation from Gordo et al. \cite{gordo17_qe} which combines it with database-side augmentation (DBA).
For diffusion, we use the implementation from Yang et al. diffusion \cite{yang2019_diffusion}.
In both cases, extensive grid searches were performed to find the best hyperparameters.
%
%
We compute results for DenseVlad \cite{torii2018_tokyo} and SuperGlue \cite{sarlin20_superglue} using the implementations provided by the respective authors.
%
%
We compare to the spatial verification approach from DELG \cite{cao20_delg}, using the official implementation trained on the Google Landmark Dataset \cite{noh17_delf}. Given the nature of the method, it cannot be fine-tuned on visual geolocalization datasets in a weakly supervised manner.
%
%
We compare to InLoc \cite{taira2018_inloc}, which uses dense local features to re-rank the predictions via RANSAC.
%
%
Lastly, since our warping regression module was inspired by Rocco et al. \cite{rocco2018_warp}, we test the dense matching strategy from {\our}, but replacing our module $W$ with the one from \cite{rocco2018_warp}. In particular, we use their released VGG16-based model trained on images from Google StreetView in Tokyo.

The results of all methods are reported in \cref{tab:results_competitors}, which shows that on average {\our} gives more accurate results than all other methods, on varying distance thresholds. 
We see that both diffusion and QE+DBA fail to achieve significant improvements.
This is likely due to the fact that visual geolocalization datasets usually have just a few positives per query, whereas these methods work best with landmark retrieval datasets, where positives are abundant.
%
%
The spatial verification with DELG \cite{cao20_delg} shows slight improvements only when using a high distance threshold (50m) for positive results. 
%
%
InLoc \cite{taira2018_inloc} proves detrimental for re-ranking NetVLAD's global features, which is likely due to its reliance of low level local features (extracted from the VGG's conv5 and conv3) which are not robust to outdoor scene variations.
%
%
Finally, the dense matching using the warping from Rocco et al. \cite{rocco2018_warp} does give a small improvement, particularly on the 10m and 25m distance thresholds. However, this improvement is, on average, 2\% less than when using our warping, confirming the importance of our pairwise warping architecture and geolocalization-specific losses.

\begin{table}[t!]
\begin{adjustbox}{width=\linewidth}
\centering
\begin{tabular}{clccccccc}
\toprule
\multicolumn{1}{c}{\multirow{2}{*}{Backbone}} & \multicolumn{1}{c}{\multirow{2}{*}{Method}} & \multicolumn{3}{c}{Pitts30k} & & \multicolumn{3}{c}{R-Tokyo}\\ \cline{3-5} \cline{7-9} 
\multicolumn{2}{c}{}                                  & 10m  & 25m  & 50m  & & 10m  & 25m  & 50m  \\ \hline
VGG16 & NV                                       & 67.2 & 82.5 & 86.5 & & 50.2 & 56.9 & 59.9 \\
VGG16 & NV + QE + DBA \cite{gordo17_qe}          & 66.5 & 82.3 & 86.4 & & 51.2 & 58.2 & 62.3 \\
VGG16 & NV + diffusion \cite{yang2019_diffusion} & 65.1 & 80.9 & 85.5 & & 52.2 & 58.6 & 62.0 \\
VGG16 & NV + Rocco et al. \cite{rocco2018_warp}  & 68.4 & 81.9 & 85.7 & & 51.9 & 59.3 & 61.0 \\
VGG16 & InLoc \cite{taira2018_inloc}                  & 44.5 & 72.8 & 79.5 & & 36.4 & 49.5 & 53.5 \\ \hline
VGG16 & NV + GeoWarp \textbf{(ours)} & \textbf{70.2} & \textbf{83.3} & \textbf{86.7} & & \textbf{54.5}& \textbf{61.6} & \textbf{63.6} \\ 
\hline \hline
- & DenseVLAD \cite{torii2018_tokyo}    & 63.6 & 77.3 & 81.6 & & 35.4 & 39.4 & 40.1 \\
- & SuperGlue \cite{sarlin20_superglue} & 72.0 & \textbf{84.9} & \textbf{88.1} & & 65.3 & 73.1 & 74.7 \\
ResNet-50 & NV                                      & 70.2 & 84.3 & 87.3 & & 65.0 & 72.4 & 74.4 \\
ResNet-50 & NV + QE + DBA \cite{gordo17_qe}         & 68.6 & 83.7 & 87.1 & & 66.7 & 72.1 & 74.1 \\
ResNet-50 & NV + diffusion \cite{yang2019_diffusion}& 67.6 & 81.6 & 85.1 & & 62.3 & 69.0 & 72.4 \\
ResNet-50 & DELG \cite{cao20_delg}                       & 65.4 & 83.0 & 88.0 & & 60.6 & 73.0 & 75.1 \\ \hline
ResNet-50 & NV + GeoWarp \textbf{(ours)} & \textbf{72.1} & 84.8 & 87.8 & &\textbf{69.7} & \textbf{74.4} & \textbf{75.4} \\ \bottomrule
\end{tabular}
\end{adjustbox}
\vspace{2pt}
\caption{Recall@1 of state-of-the-art methods for geolocalization and retrieval. NV stands for NetVLAD \cite{arandjelovic2016_netvlad}. Note that DenseVLAD does not use a CNN backbone, while SuperGlue uses one ad hoc.}
\label{tab:results_competitors}
\end{table}


\subsection{Ablation study}
\noindent
We perform an extensive ablation study (see \cref{tab:ablation_losses}) to verify the significance of each term in the total loss \cref{eq:total_loss}. 
We see that without the guidance of the self-supervised loss \cref{eq:loss_ss}, re-ranking with local features is generally detrimental with respect to the baseline.
As expected, using solely the consistency loss gives results on par with not training the homography regression, as the warping simply performs an identity transformation.
While the effect of separately applying each of the two weakly supervised losses leads to similar results, the orthogonality of each loss's gains is clear, as the three losses combined achieve best results.

\begin{figure}
    \centering
    \includegraphics[width=0.8\columnwidth]{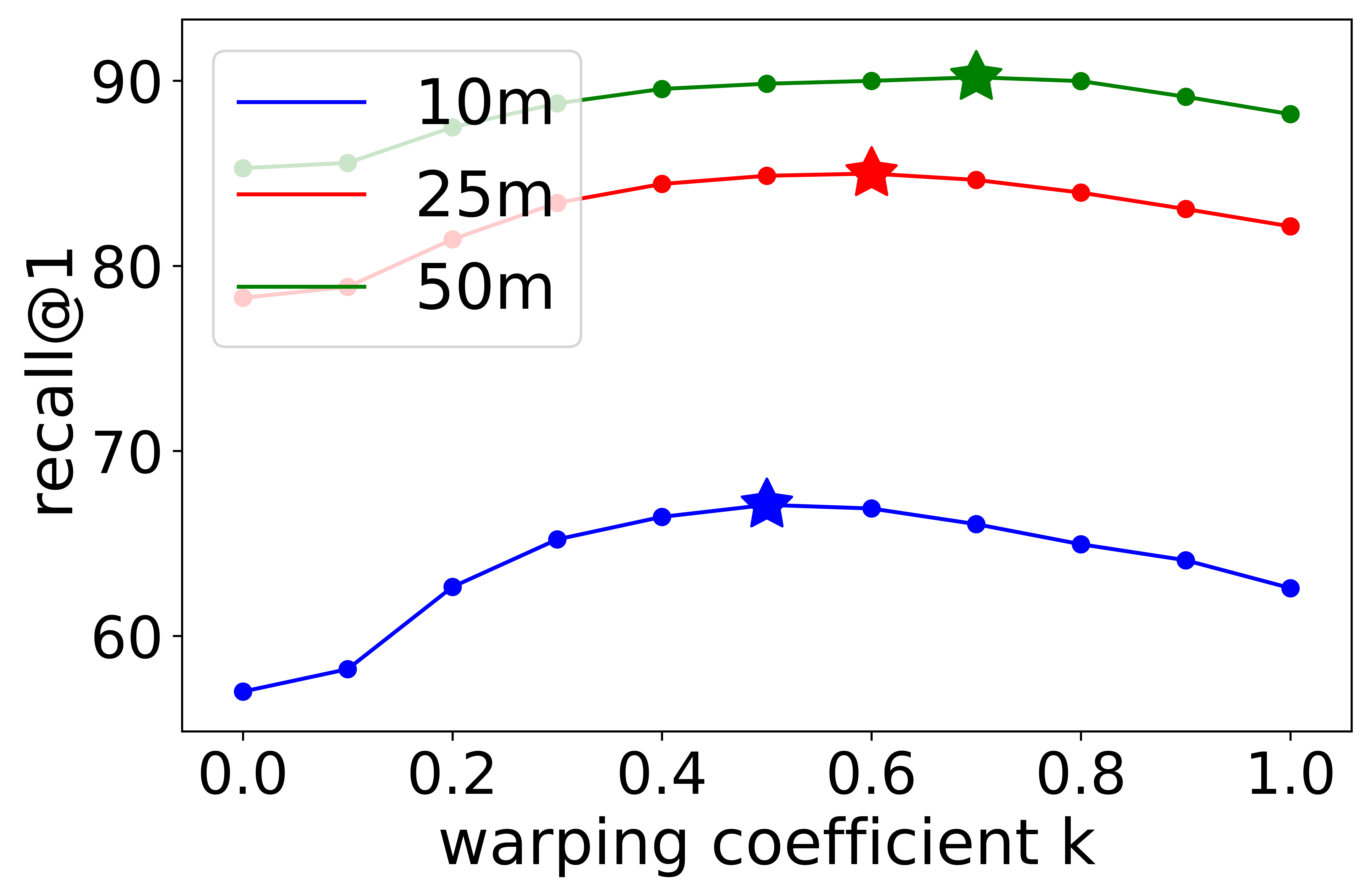}
    \vspace{2pt}
    \caption{Ablation of warping coefficient $k$. Higher $k$ correspond to more aggressive warping, which is useful when a rougher estimate of the position is needed.}
    \label{fig:ablation_k}
\end{figure}

We also perform a second ablation study focused on the parameter $k$ used in the self-supervised training method. 
\Cref{fig:ablation_k} shows the recall@1 curves at different values of $k$. 
We observe that higher values of $k$ are better for coarser localization problems. This is due to the fact that a high $k$ produces more difficult training images, inducing the warping regression module to learn more aggressive homography transformations. 
On the other hand, when the goal is to achieve a more precise localization it is undesirable to warp too much the query-prediction pairs, and a lower $k$ is more appropriate.
Qualitative results of the effect of the parameter $k$ are reported in the supplementary material.

\begin{table}[t!]
\begin{adjustbox}{width=\linewidth}
\centering
\begin{tabular}{cccc|cccc}
\toprule
\begin{tabular}[c]{@{}c@{}}Features\\ type\end{tabular} & $\lambda_{ss}$ & $\lambda_{fw}$ & $\lambda_{cons}$ & \begin{tabular}[c]{@{}c@{}}VGG16\\ GeM\end{tabular} & \begin{tabular}[c]{@{}c@{}}VGG16\\ NV\end{tabular} & \begin{tabular}[c]{@{}c@{}}ResNet50\\ GeM\end{tabular} & \begin{tabular}[c]{@{}c@{}}ResNet50\\ NV\end{tabular} \\ \hline
global & 0 & 0  & 0   & 70.5 & 85.1 & 82.9 & 86.4 \\ 
local  & 0 & 0  & 0   & 74.3 & 83.2 & 78.2 & 80.7 \\ 
local  & 0 & 0  & 0.1 & 74.2 & 83.1 & 78.3 & 80.8 \\
local  & 0 & 10 & 0   & 74.3 & 82.9 & 78.1 & 80.9 \\
local  & 0 & 10 & 0.1 & 74.3 & 83.0 & 77.8 & 80.8 \\
local  & 1 & 0  & 0   & 78.6 & 87.0 & 84.2 & 87.0 \\ 
local  & 1 & 0  & 0.1 & 78.8 & 87.1 & 84.3 & 86.8 \\
local  & 1 & 10 & 0   & 78.8 & 87.1 & \textbf{84.6} & 86.9 \\
local  & 1 & 10 & 0.1 & \textbf{79.0} & \textbf{87.2} & 84.5 & \textbf{87.2} \\ \bottomrule
\end{tabular}
\end{adjustbox}
\vspace{2pt}
\caption{Ablation results over the three losses on the validation set of Pitts30k \cite{arandjelovic2016_netvlad}.}
\label{tab:ablation_losses}
\end{table}

\section{Conclusions}

\noindent
We have presented {\our}, a novel method for image matching tailored for visual geolocalization. {\our} combines the robustness of deep dense local features to clutter and lighting variations with a learnable architecture that provides invariance to viewpoint shifts. This novel architecture is trained with novel loss functions that only require unlabeled data or can take advantage of weak labels.
We evaluate {\our} on two well-established datasets, demonstrating the soundness of our approach against state-of-the-art retrieval and geolocalization methods, over which we show significant improvements. 

\paragraph{Acknowledgements}
We acknowledge that the research activity herein was carried out using the IIT HPC infrastructure.

{\small
\bibliographystyle{ieee_fullname}
\bibliography{main}
}

\section{Appendix}

\section{Qualitative results}
\begin{figure*}
    \centering
\resizebox{\textwidth}{!}{
\includegraphics[height=5cm]{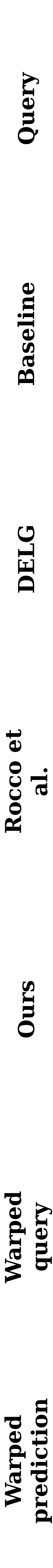}
\includegraphics[height=5cm]{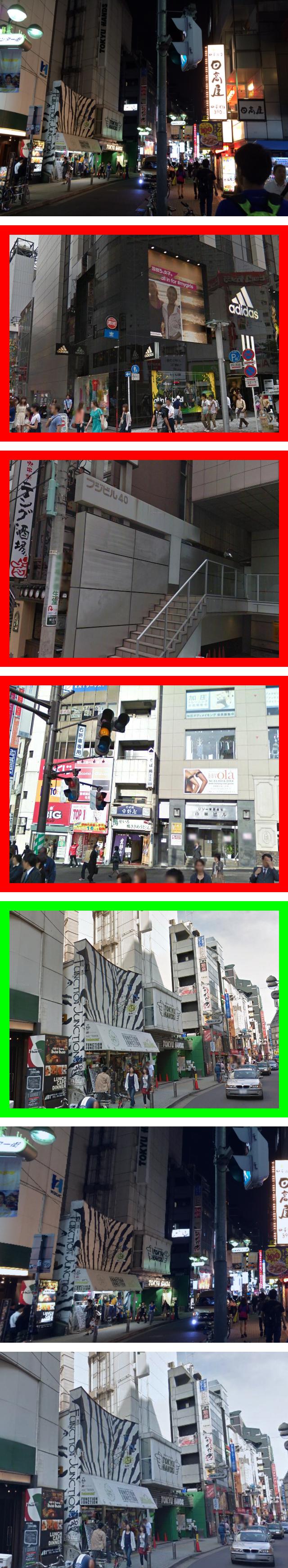}
\includegraphics[height=5cm]{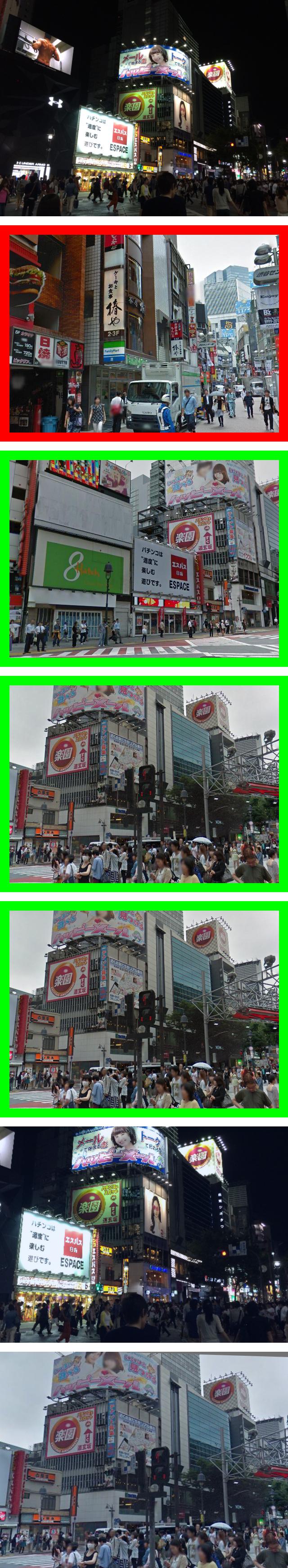}
\includegraphics[height=5cm]{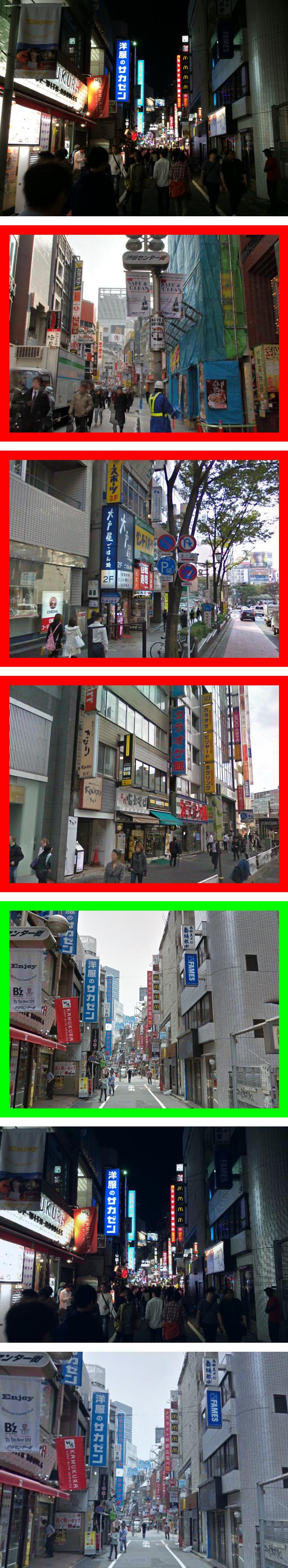}
\includegraphics[height=5cm]{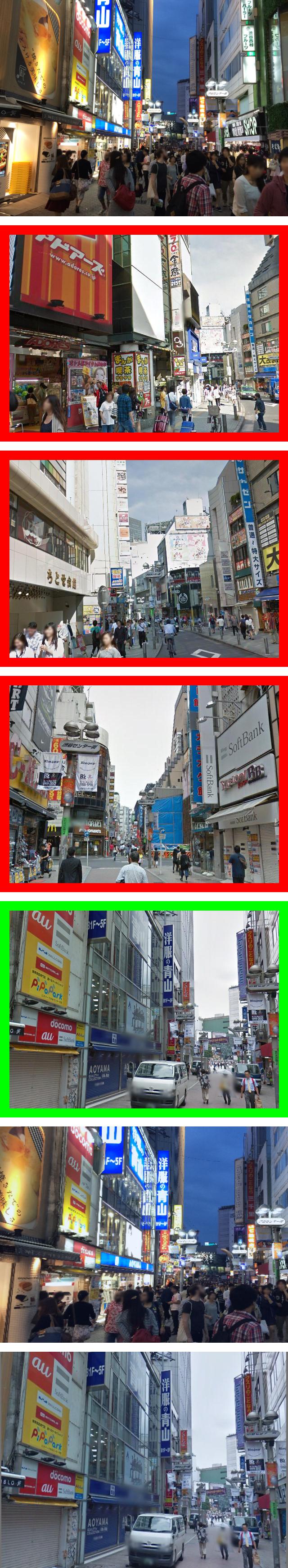}
\includegraphics[height=5cm]{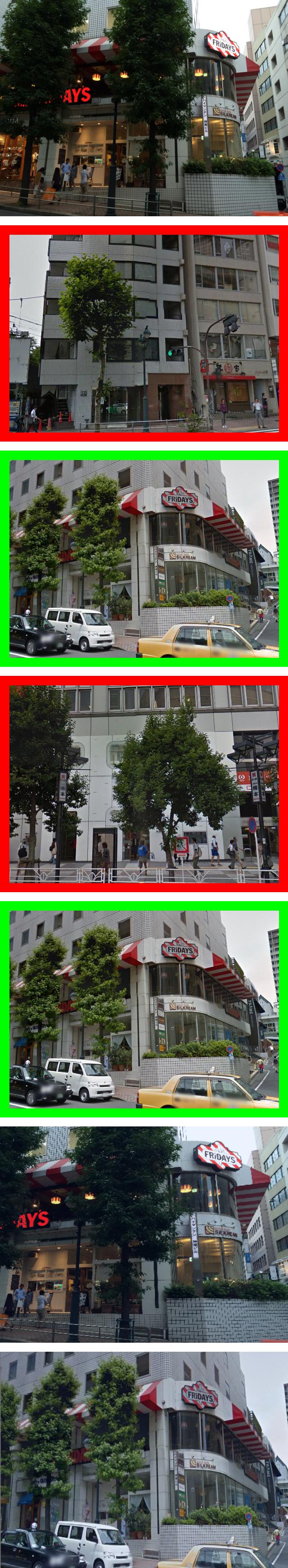}
\includegraphics[height=5cm]{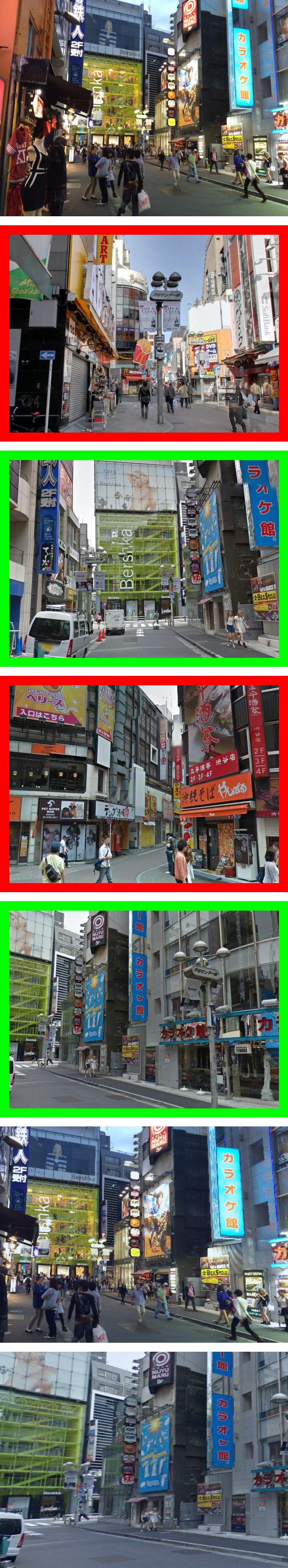}
\includegraphics[height=5cm]{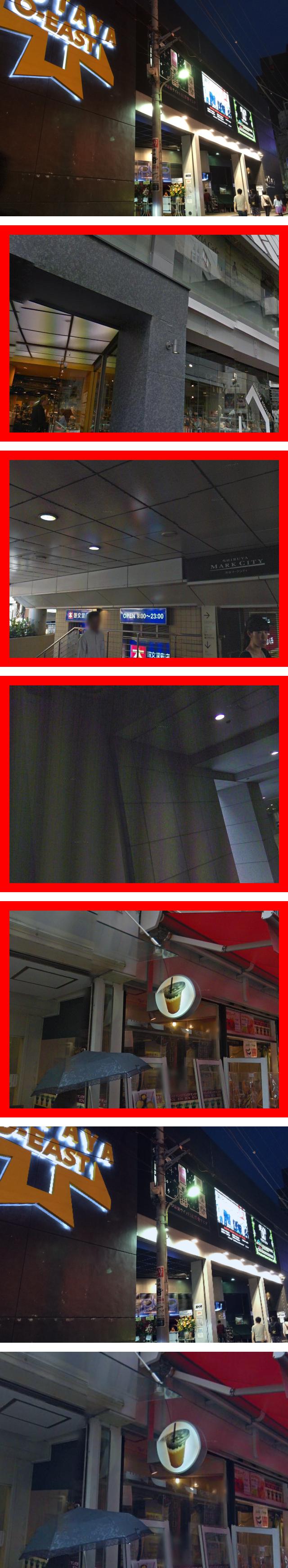}
}
\caption{Qualitative results, each column corresponds to one test case. The first row represents the queries, the next four rows show the first prediction with various methods, and the last two rows show the warped queries and first prediction with GeoWarp (\textbf{Ours}).}
\label{fig:comparisons}
\end{figure*}

Figure \ref{fig:comparisons} displays qualitative results that extend the quantitative discussion of the experiments presented in the main paper.
The figure showcases the results achieved on a few queries from the R-Tokyo \cite{warburg2020_msls} dataset. The selected queries are a good example of challenging conditions for urban visual geolocalization, as they are images taken at night/evening (whereas the gallery is built from images captured during the day) and with many dynamic and static objects occluding the scene in the background.
In particular, we show the top-1 retrieved result from  a shortlist of 5 predictions. 
We note that the best baseline (ResNet-50 \cite{he2016_resnet} + NetVLAD \cite{arandjelovic2016_netvlad}) fails in all the examples.
Spatial verification using the method from DELG \cite{cao20_delg} manages to correctly localize only a few of the queries. 
Lastly, building our proposed dense matching method but using the warping module from \cite{rocco2018_warp}, which is not tailored for the geolocalization task, only manages to retrieve a correct prediction for one of the queries, particularly when the query and the prediction have only small perspective difference and a large overlap.
On the other hand, GeoWarp manages to correctly localize all but one query. However, this failure case is due to the fact that the top-5 predictions retrieved by the global search on the gallery do not contain any positive match for the query, therefore re-ranking them cannot help improving the localization.
The last two rows from Fig. \ref{fig:comparisons} illustrate the warped query and the first prediction generated with GeoWarp.

\section{Effects of $k$ on the warping operation}
\begin{figure*}
    \centering
\resizebox{\textwidth}{!}{
\includegraphics[height=2cm]{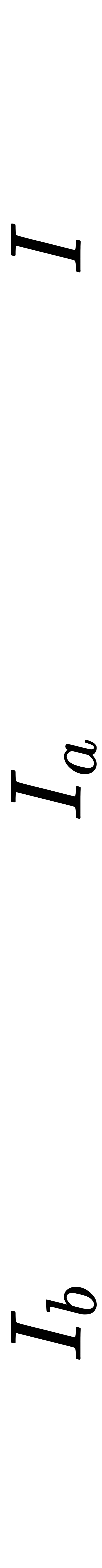}
\includegraphics[height=2cm]{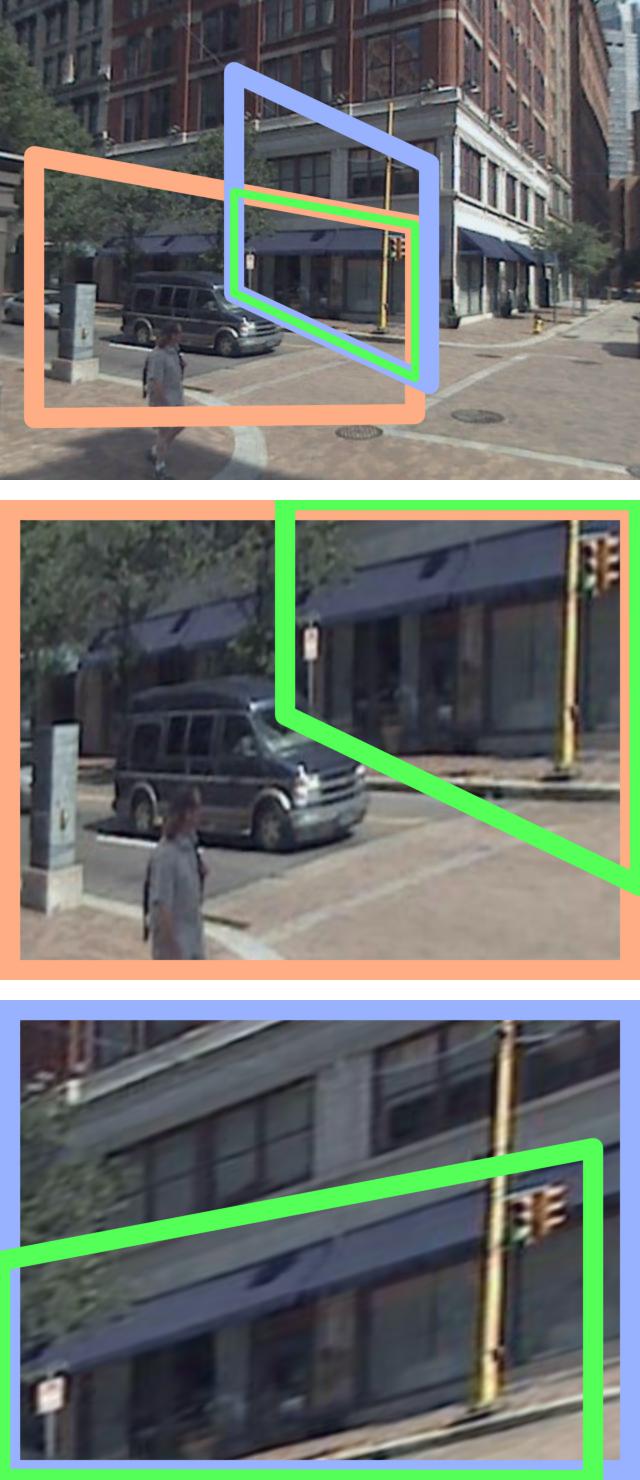}
\includegraphics[height=2cm]{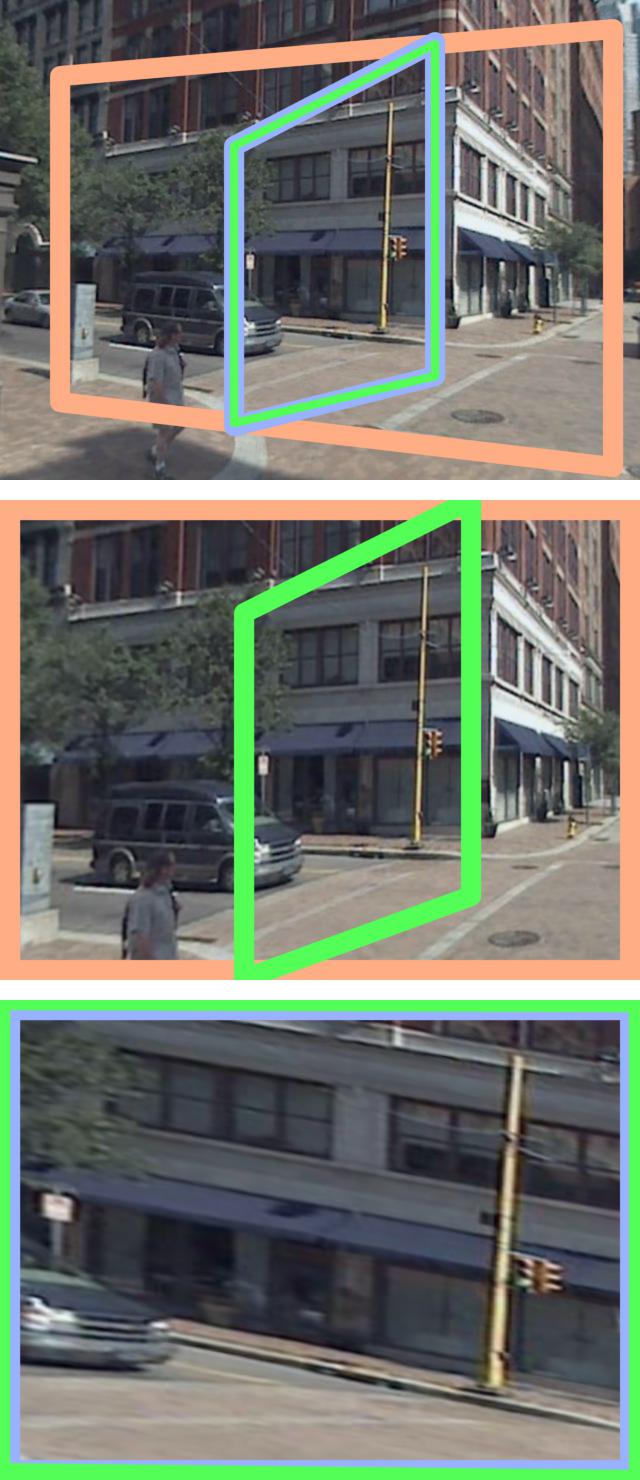}
\includegraphics[height=2cm]{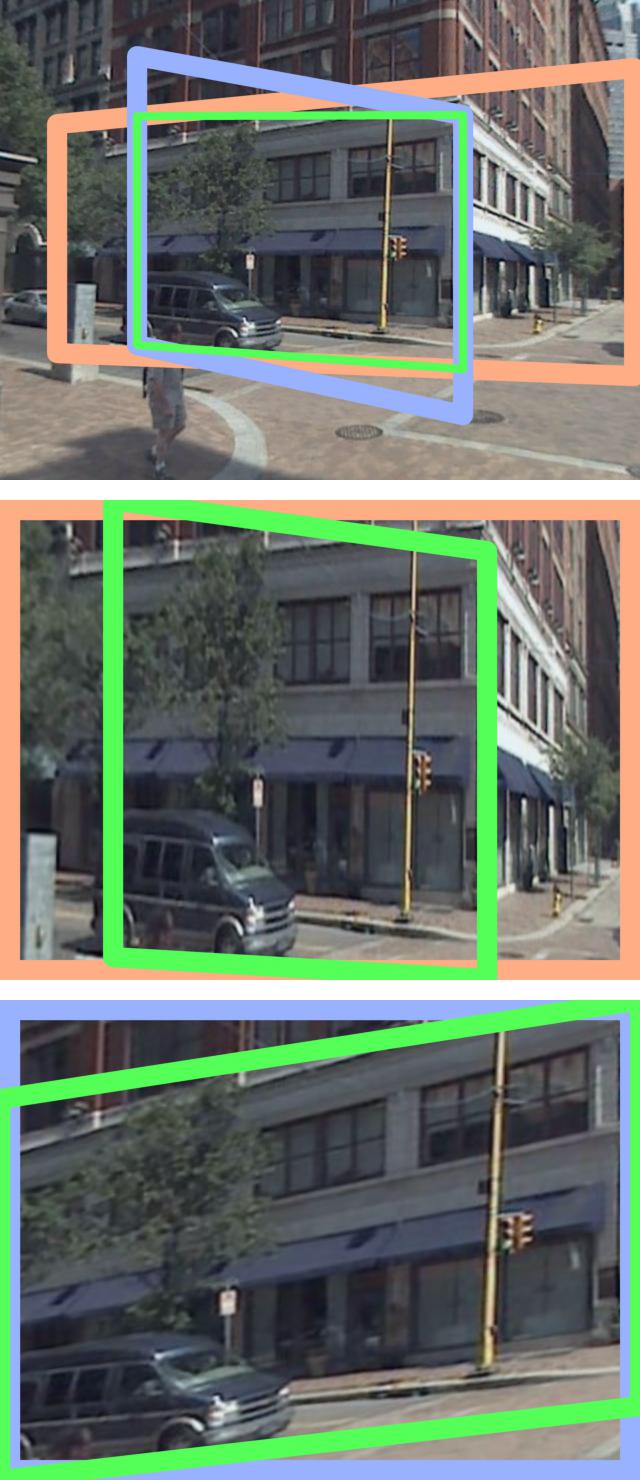}
\includegraphics[height=2cm]{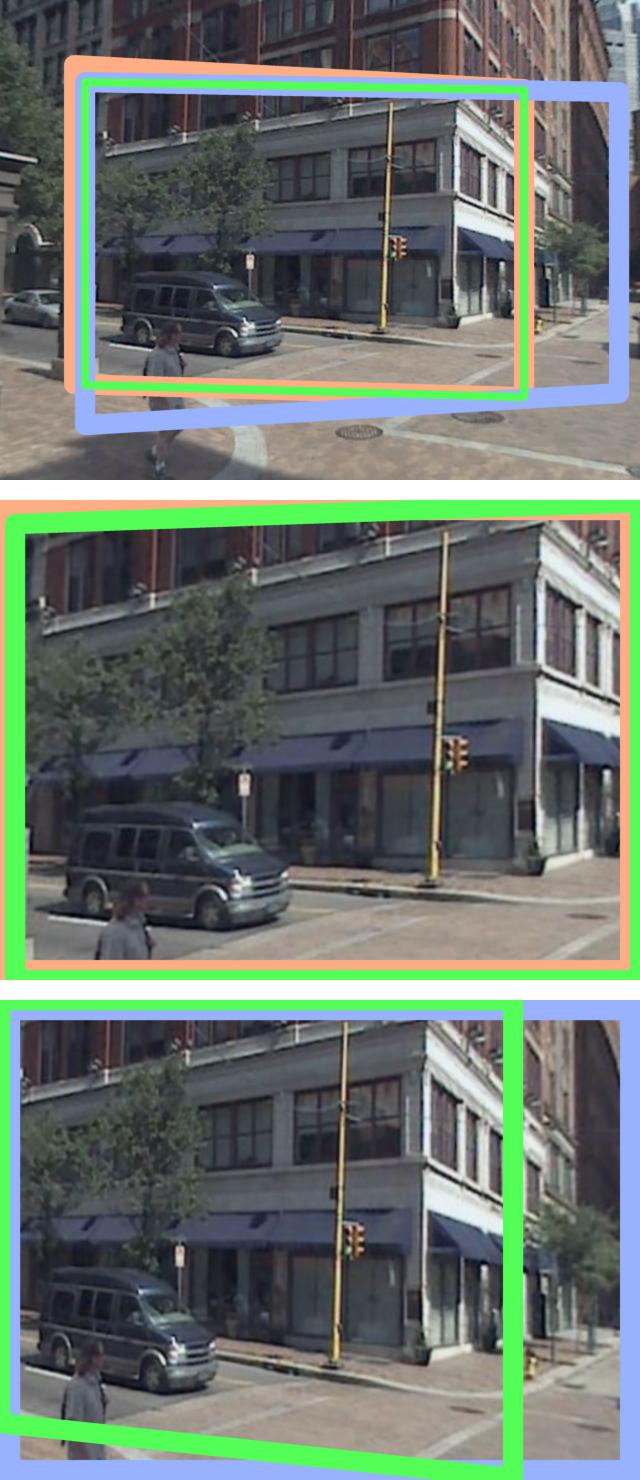}
\includegraphics[height=2cm]{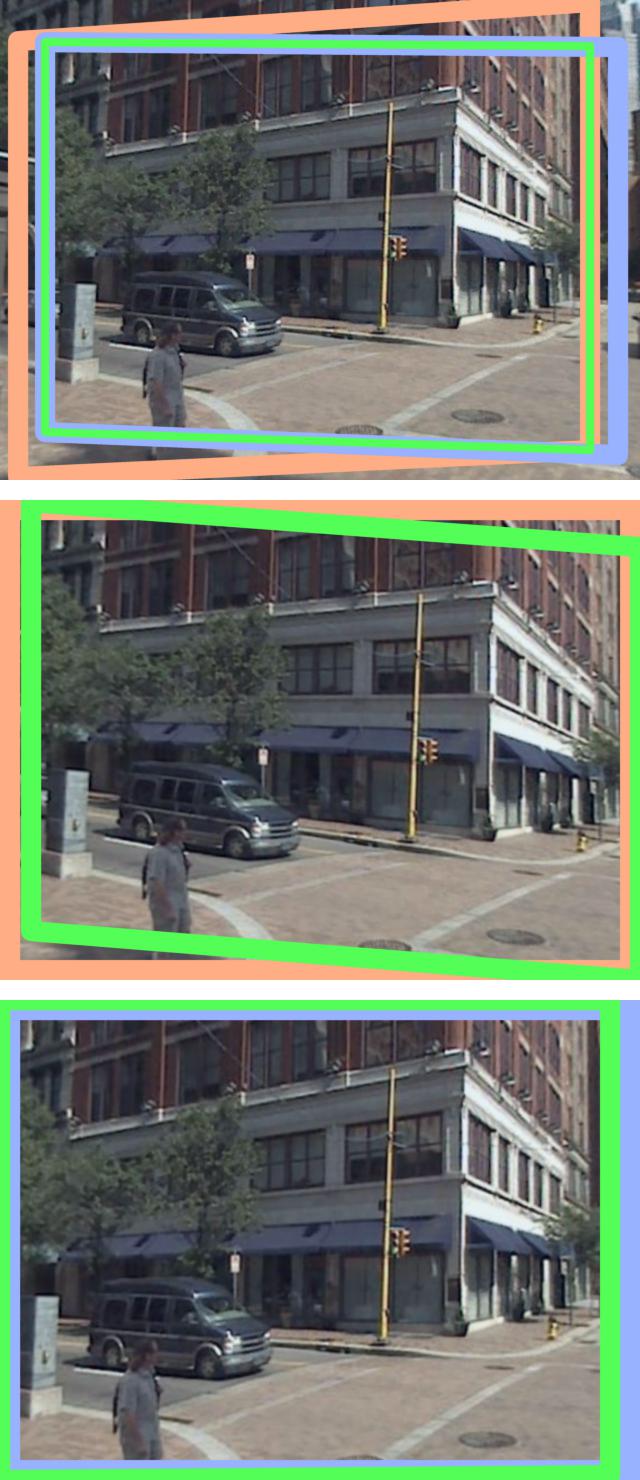}
}
\resizebox{\textwidth}{!}{\includegraphics{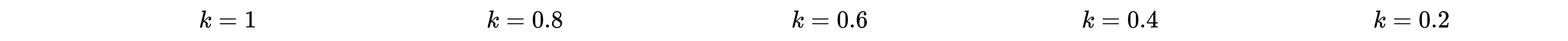}}
\caption{Examples of training quadruplets generated in a self-supervised fashion, using different values of $k$ (1, 0.8, 0.6, 0.4, 0.2). For each column, the top image represents the source image $I$, the two other images represent $I_a$ and $I_b$. The green quadrilaterals in the second and third rows indicate $\boldsymbol{t}_a$ and $\boldsymbol{t}_b$, respectively.}
\label{fig:ss_data}
\end{figure*}

The main paper includes a quantitative ablation study on the impact of $k \in [0, 1]$ on the recall@1 metric.
Here, we give further intuition about the influence of $k$ on the self-supervised generation of the training quadruplets $\{I_a, I_b, \boldsymbol{t}_a, \boldsymbol{t}_b\}$.
In Fig. \ref{fig:ss_data} we see that for small values of $k$ the training quadruplets are generated sampling quadrilaterals whose corners are close to the corners of the image $I$. This means that the two images $I_a$ and $I_b$ have large overlaps and little perspective difference.
On the other hand, high values of $k$ lead to the images $I_a$ and $I_b$ to have have little overlap, simulating very different views of the same scene.

\begin{figure*}
    \centering
    \parbox[b]{.02\linewidth}{\subcaption{}}
    \begin{minipage}{.18\textwidth}
        \begin{subfigure}{\textwidth}
            \includegraphics[width=\textwidth]{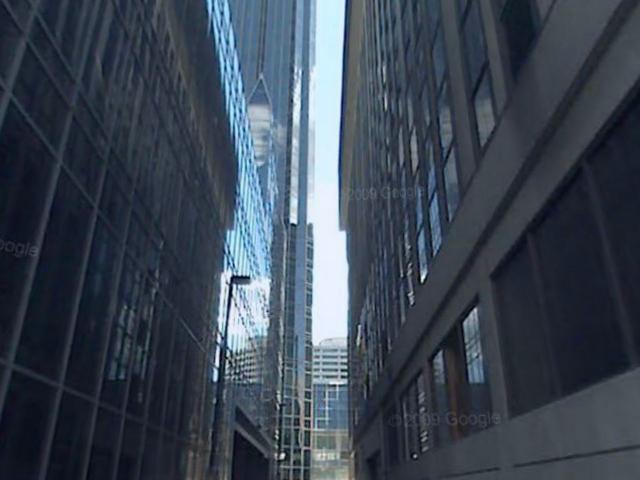}
        \end{subfigure}
        \begin{subfigure}{\textwidth}
            \includegraphics[width=\textwidth]{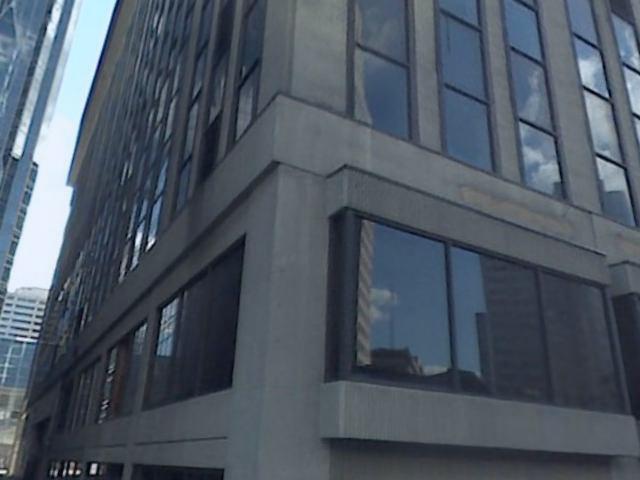} 
        \end{subfigure}
    \end{minipage}
    \hspace{2mm}\vline\hspace{2mm}
    \begin{minipage}{.18\textwidth}
        \begin{subfigure}{\textwidth}
            \includegraphics[width=\textwidth]{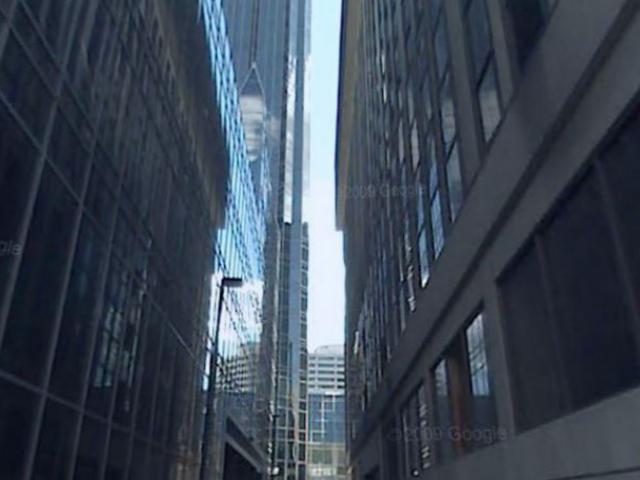}
        \end{subfigure}
        \begin{subfigure}{\textwidth}
            \includegraphics[width=\textwidth]{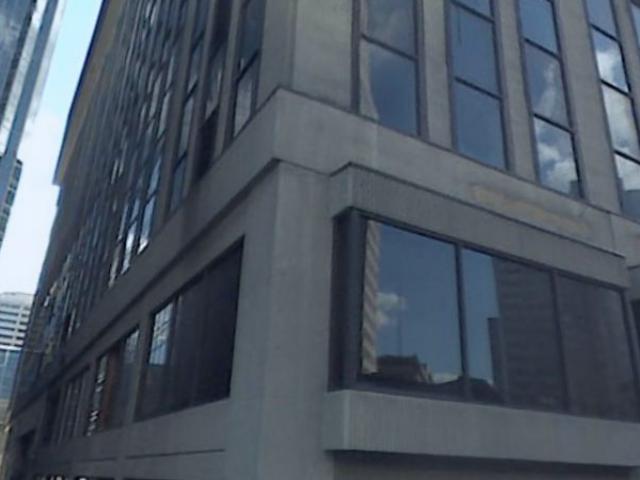} 
        \end{subfigure}
    \end{minipage}
    \begin{minipage}{.18\textwidth}
        \begin{subfigure}{\textwidth}
            \includegraphics[width=\textwidth]{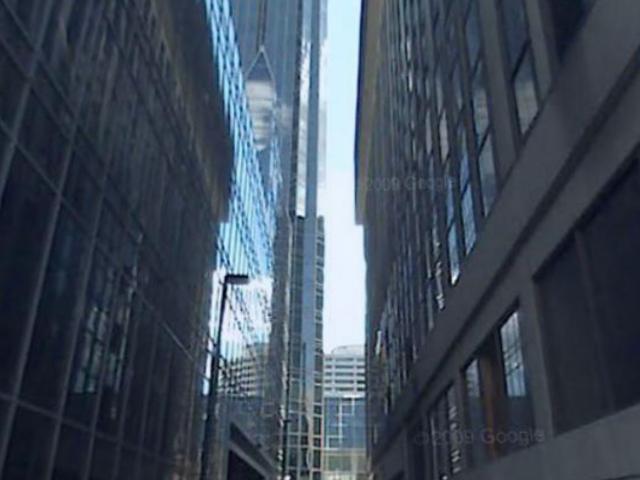}
        \end{subfigure}
        \begin{subfigure}{\textwidth}
            \includegraphics[width=\textwidth]{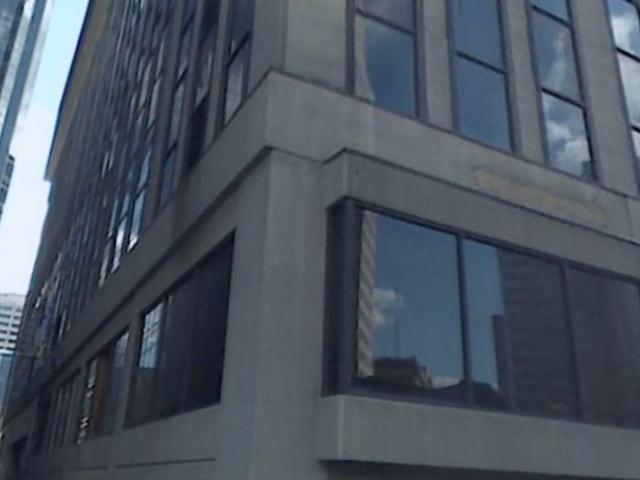} 
        \end{subfigure}
    \end{minipage}
    \begin{minipage}{.18\textwidth}
        \begin{subfigure}{\textwidth}
            \includegraphics[width=\textwidth]{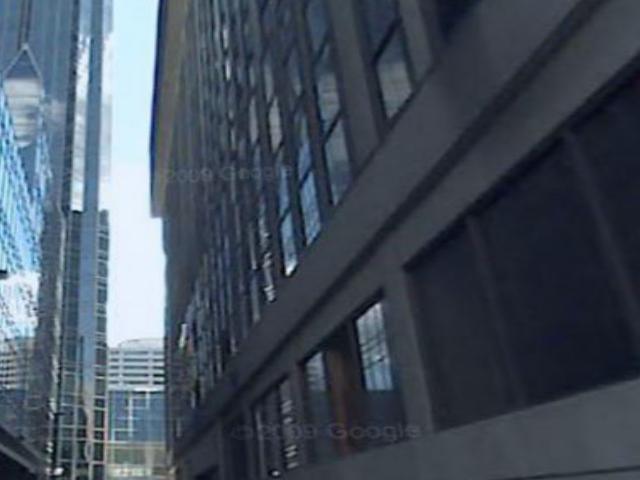}
        \end{subfigure}
        \begin{subfigure}{\textwidth}
            \includegraphics[width=\textwidth]{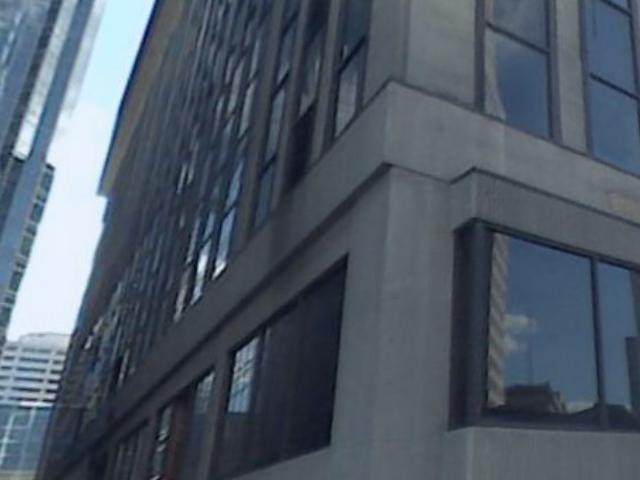} 
        \end{subfigure}
    \end{minipage}
    \begin{minipage}{.18\textwidth}
        \begin{subfigure}{\textwidth}
            \includegraphics[width=\textwidth]{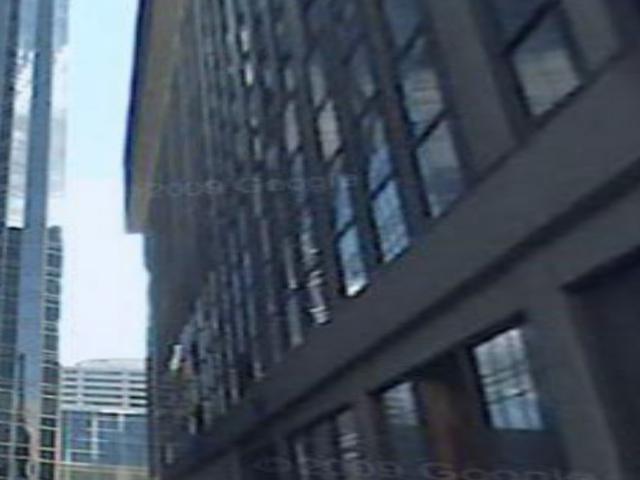}
        \end{subfigure}
        \begin{subfigure}{\textwidth}
            \includegraphics[width=\textwidth]{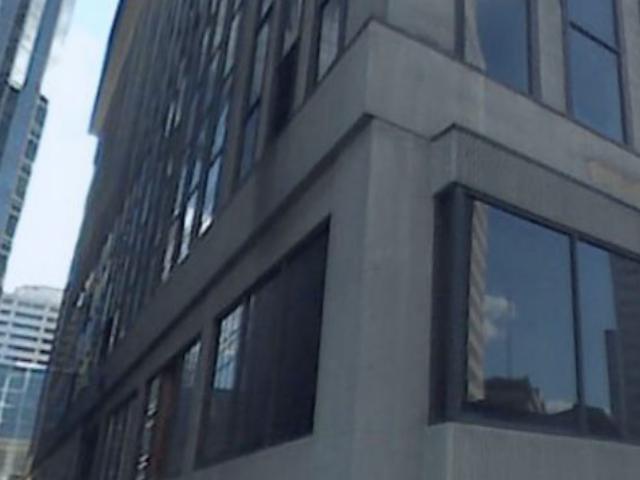} 
        \end{subfigure}
    \end{minipage}
    
    \vspace{3mm}
    \parbox[b]{.02\linewidth}{\subcaption{}}
    \begin{minipage}{.18\textwidth}
        \begin{subfigure}{\textwidth}
            \includegraphics[width=\textwidth]{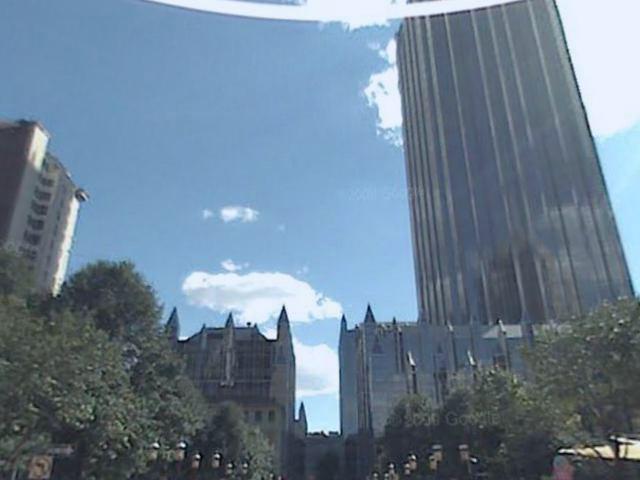}
        \end{subfigure}
        \begin{subfigure}{\textwidth}
            \includegraphics[width=\textwidth]{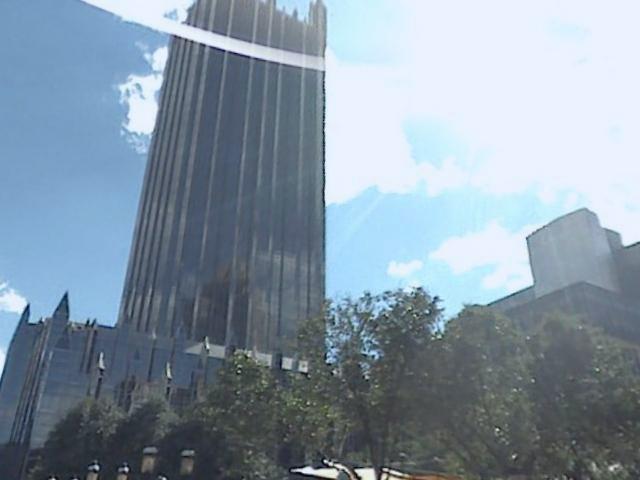} 
        \end{subfigure}
    \end{minipage}
    \hspace{2mm}\vline\hspace{2mm}
    \begin{minipage}{.18\textwidth}
        \begin{subfigure}{\textwidth}
            \includegraphics[width=\textwidth]{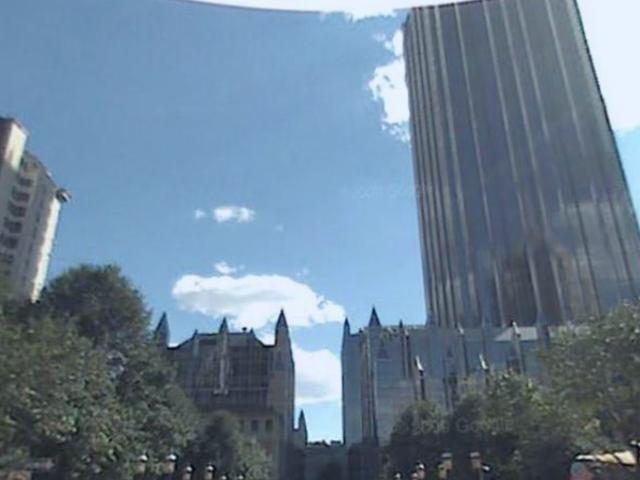}
        \end{subfigure}
        \begin{subfigure}{\textwidth}
            \includegraphics[width=\textwidth]{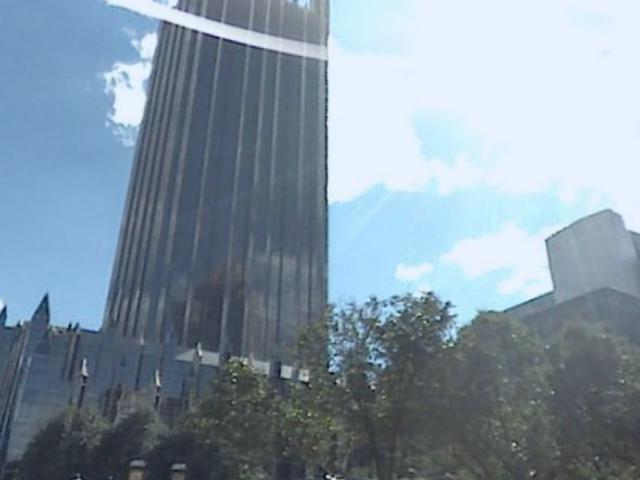} 
        \end{subfigure}
    \end{minipage}
    \begin{minipage}{.18\textwidth}
        \begin{subfigure}{\textwidth}
            \includegraphics[width=\textwidth]{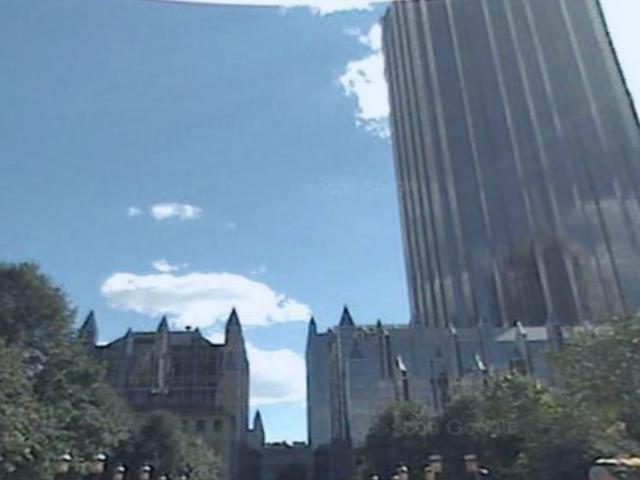}
        \end{subfigure}
        \begin{subfigure}{\textwidth}
            \includegraphics[width=\textwidth]{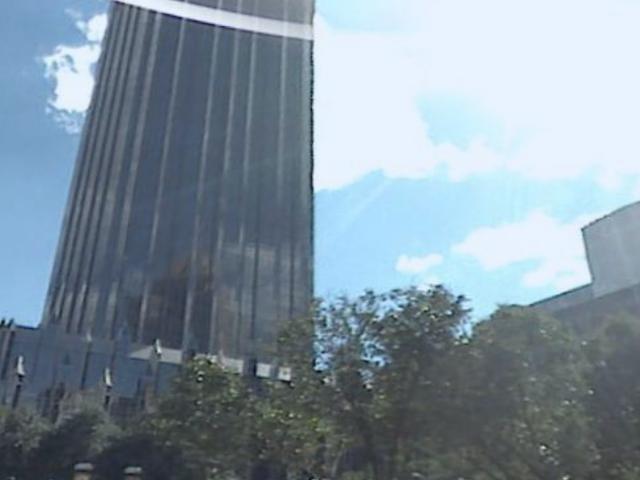} 
        \end{subfigure}
    \end{minipage}
    \begin{minipage}{.18\textwidth}
        \begin{subfigure}{\textwidth}
            \includegraphics[width=\textwidth]{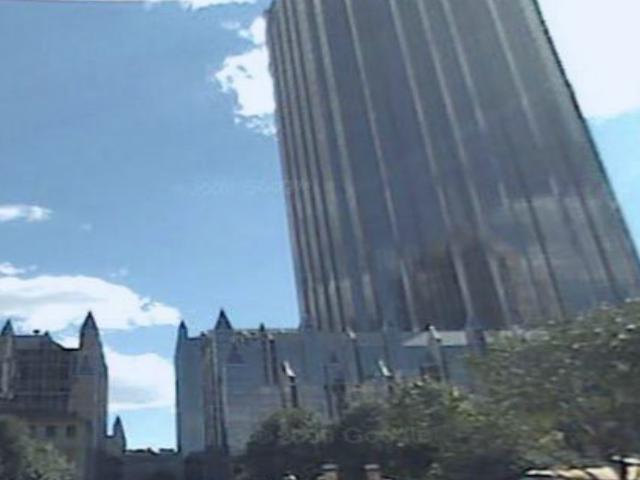}
        \end{subfigure}
        \begin{subfigure}{\textwidth}
            \includegraphics[width=\textwidth]{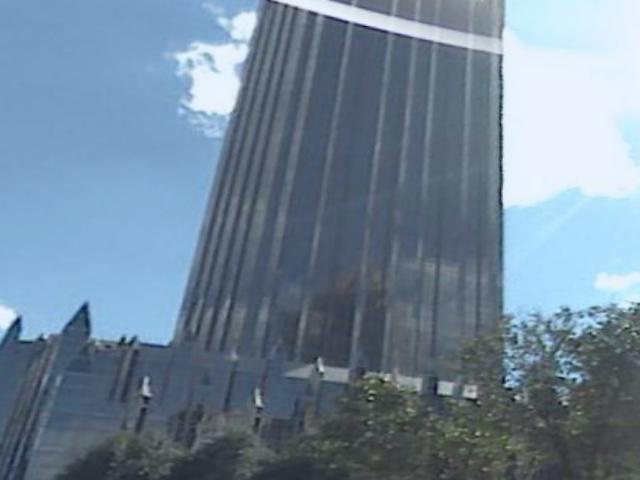} 
        \end{subfigure}
    \end{minipage}
    \begin{minipage}{.18\textwidth}
        \begin{subfigure}{\textwidth}
            \includegraphics[width=\textwidth]{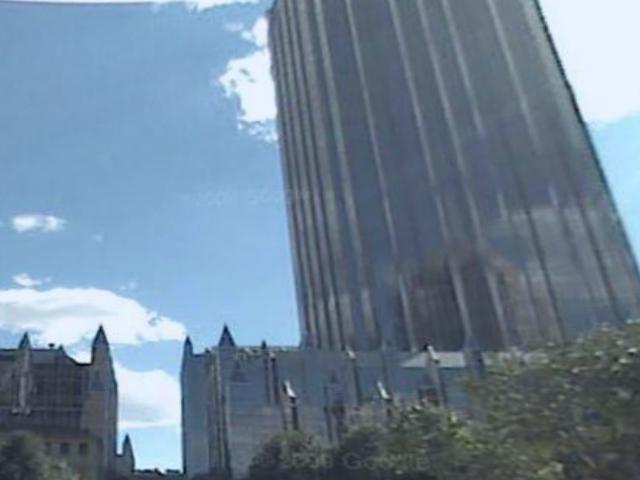}
        \end{subfigure}
        \begin{subfigure}{\textwidth}
            \includegraphics[width=\textwidth]{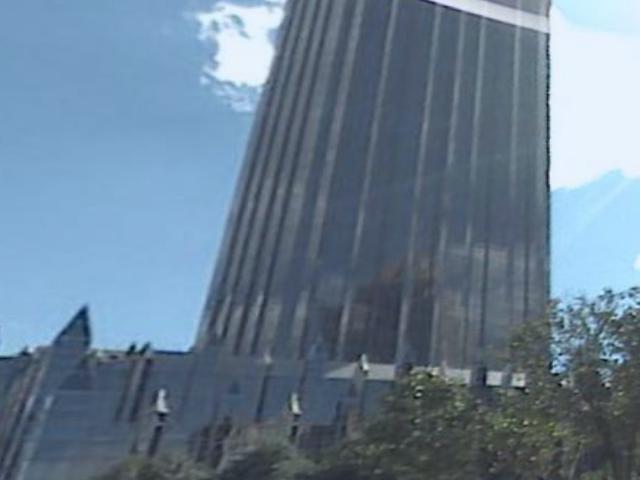} 
        \end{subfigure}
    \end{minipage}
    
    \vspace{3mm}
    \parbox[b]{.02\linewidth}{\subcaption{}}
    \begin{minipage}{.18\textwidth}
        \begin{subfigure}{\textwidth}
            \includegraphics[width=\textwidth]{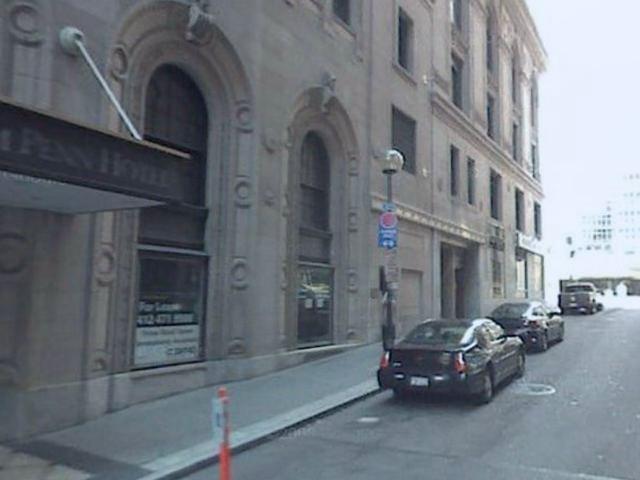}
        \end{subfigure}
        \begin{subfigure}{\textwidth}
            \includegraphics[width=\textwidth]{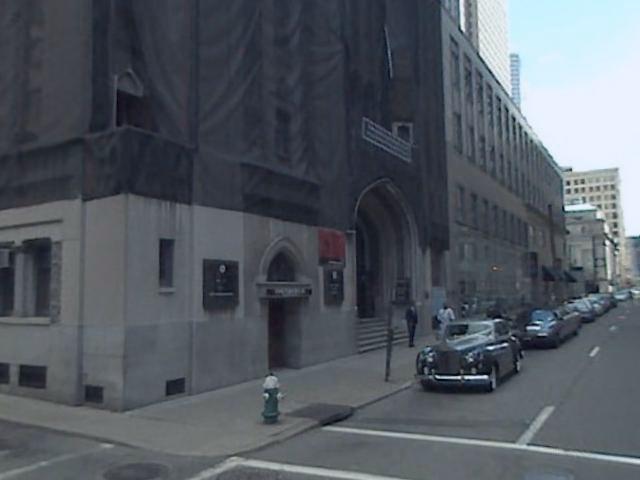} 
        \end{subfigure}
    \end{minipage}
    \hspace{2mm}\vline\hspace{2mm}
    \begin{minipage}{.18\textwidth}
        \begin{subfigure}{\textwidth}
            \includegraphics[width=\textwidth]{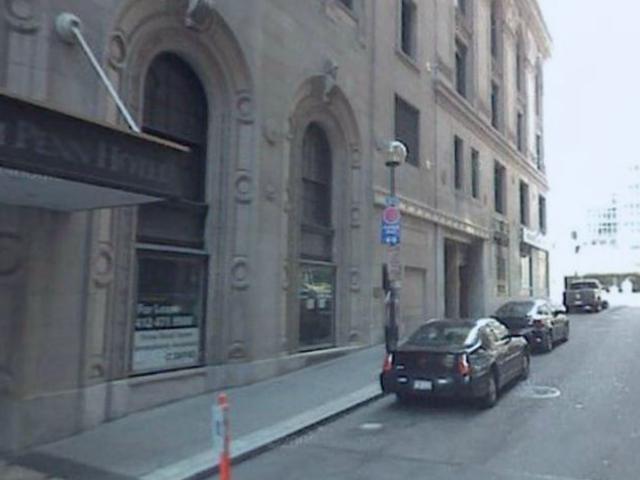}
        \end{subfigure}
        \begin{subfigure}{\textwidth}
            \includegraphics[width=\textwidth]{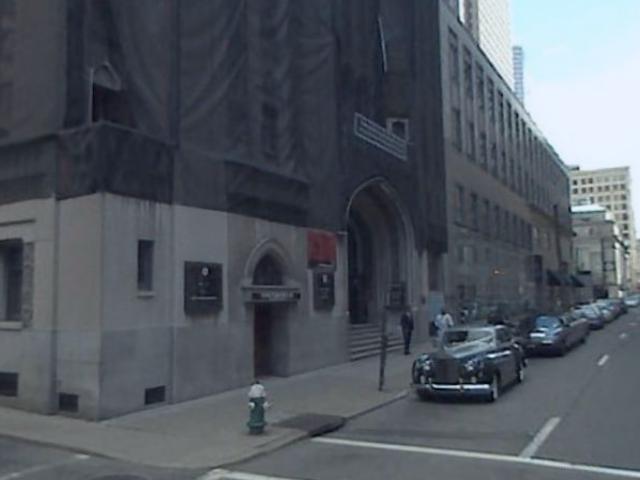} 
        \end{subfigure}
    \end{minipage}
    \begin{minipage}{.18\textwidth}
        \begin{subfigure}{\textwidth}
            \includegraphics[width=\textwidth]{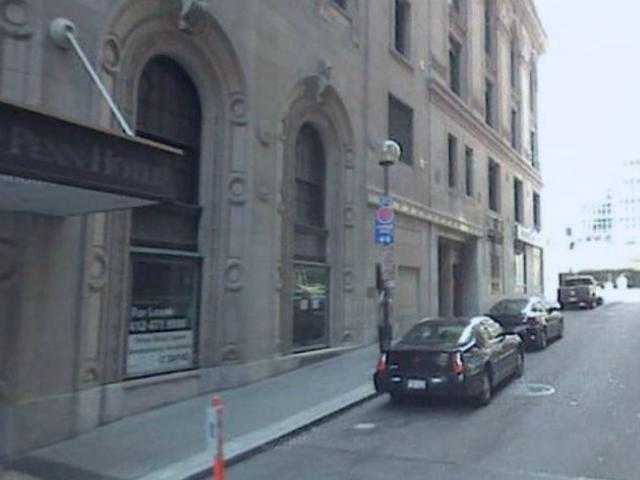}
        \end{subfigure}
        \begin{subfigure}{\textwidth}
            \includegraphics[width=\textwidth]{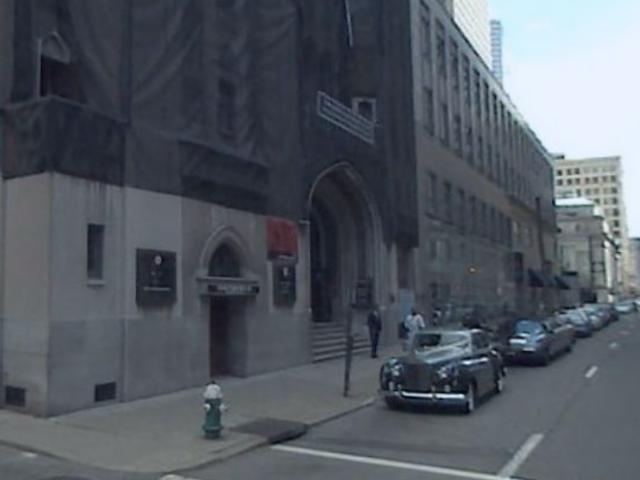} 
        \end{subfigure}
    \end{minipage}
    \begin{minipage}{.18\textwidth}
        \begin{subfigure}{\textwidth}
            \includegraphics[width=\textwidth]{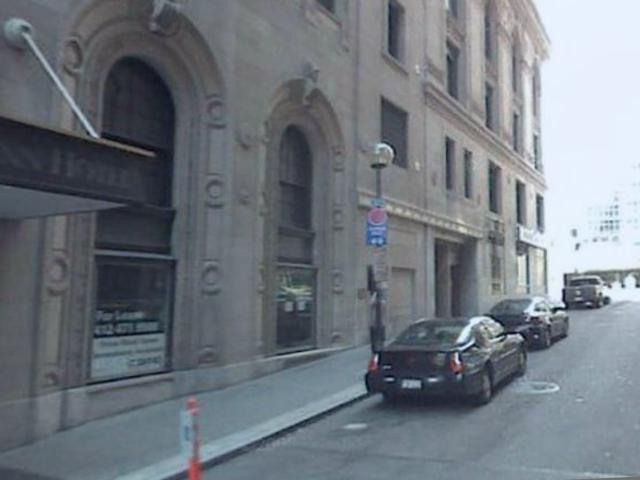}
        \end{subfigure}
        \begin{subfigure}{\textwidth}
            \includegraphics[width=\textwidth]{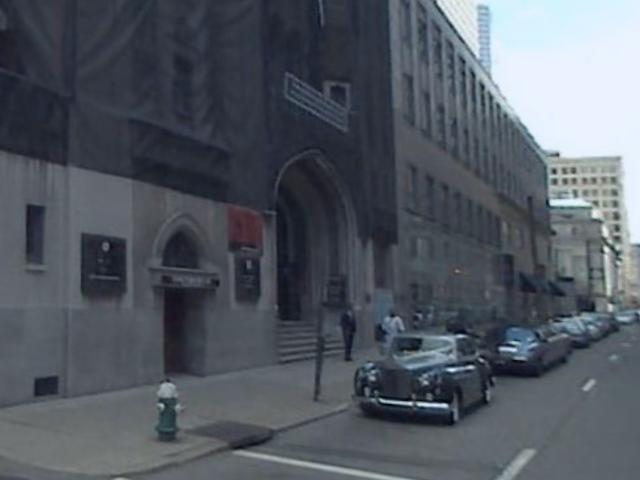} 
        \end{subfigure}
    \end{minipage}
    \begin{minipage}{.18\textwidth}
        \begin{subfigure}{\textwidth}
            \includegraphics[width=\textwidth]{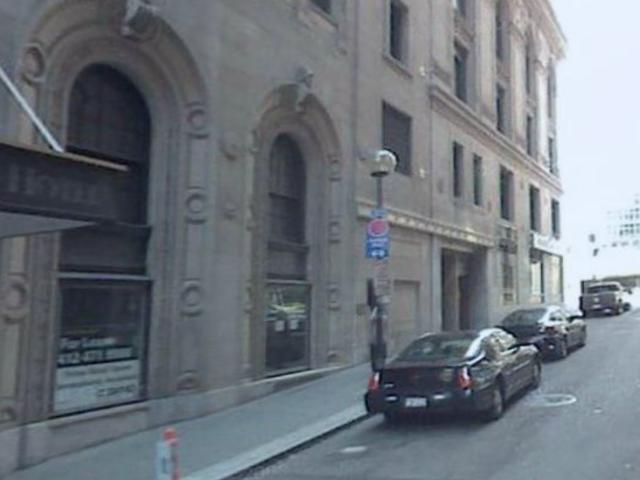}
        \end{subfigure}
        \begin{subfigure}{\textwidth}
            \includegraphics[width=\textwidth]{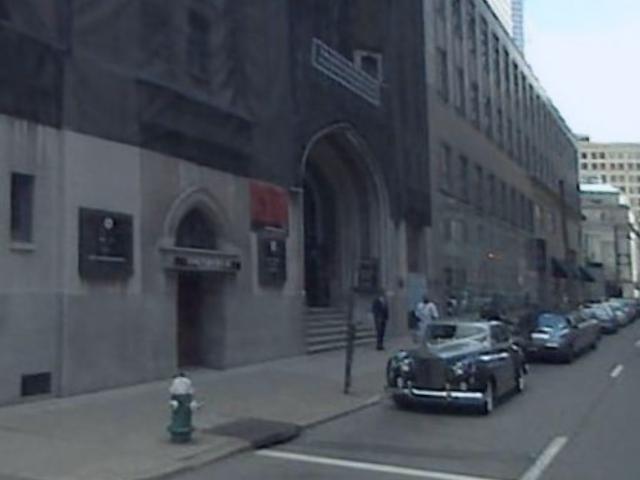} 
        \end{subfigure}
    \end{minipage}
    
    \vspace{3mm}
    \parbox[b]{.02\linewidth}{\subcaption{}}
    \begin{minipage}{.18\textwidth}
        \begin{subfigure}{\textwidth}
            \includegraphics[width=\textwidth]{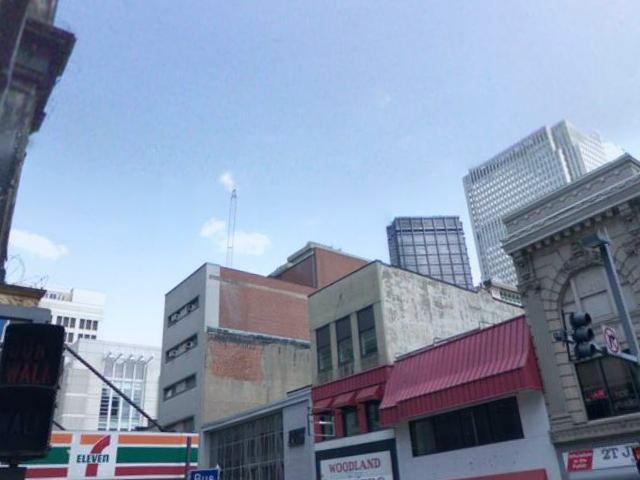}
        \end{subfigure}
        \begin{subfigure}{\textwidth}
            \includegraphics[width=\textwidth]{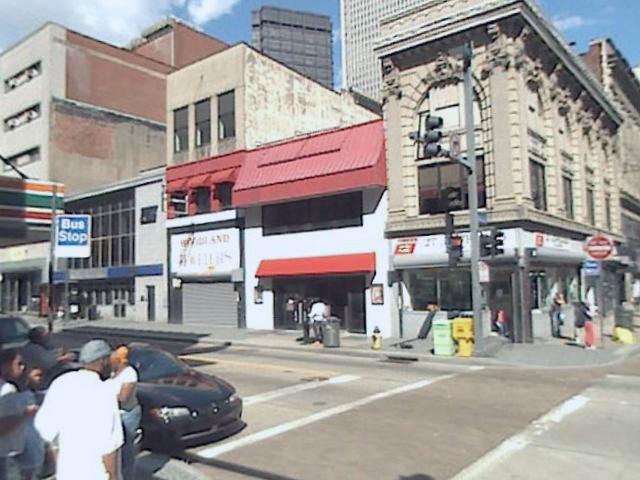} 
        \end{subfigure}
    \end{minipage}
    \hspace{2mm}\vline\hspace{2mm}
    \begin{minipage}{.18\textwidth}
        \begin{subfigure}{\textwidth}
            \includegraphics[width=\textwidth]{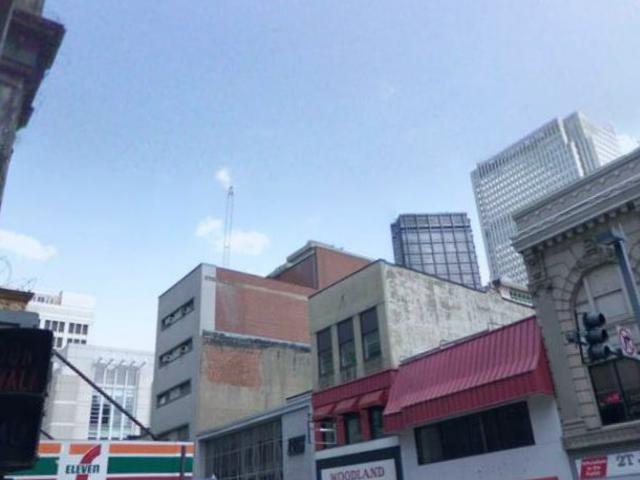}
        \end{subfigure}
        \begin{subfigure}{\textwidth}
            \includegraphics[width=\textwidth]{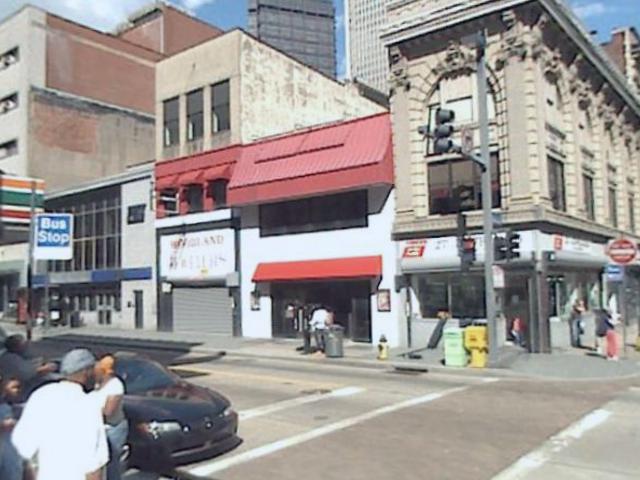} 
        \end{subfigure}
    \end{minipage}
    \begin{minipage}{.18\textwidth}
        \begin{subfigure}{\textwidth}
            \includegraphics[width=\textwidth]{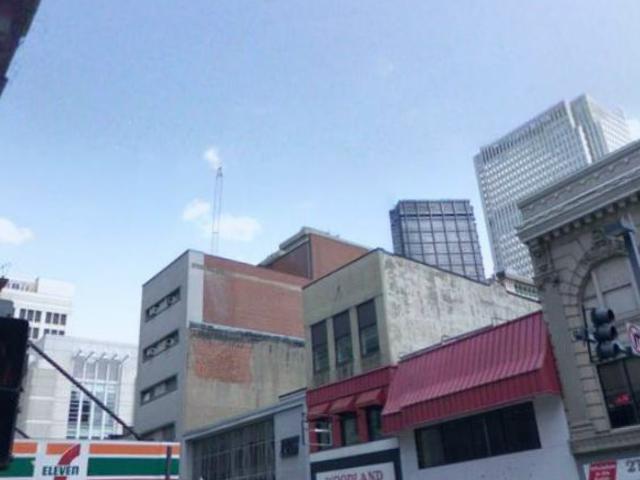}
        \end{subfigure}
        \begin{subfigure}{\textwidth}
            \includegraphics[width=\textwidth]{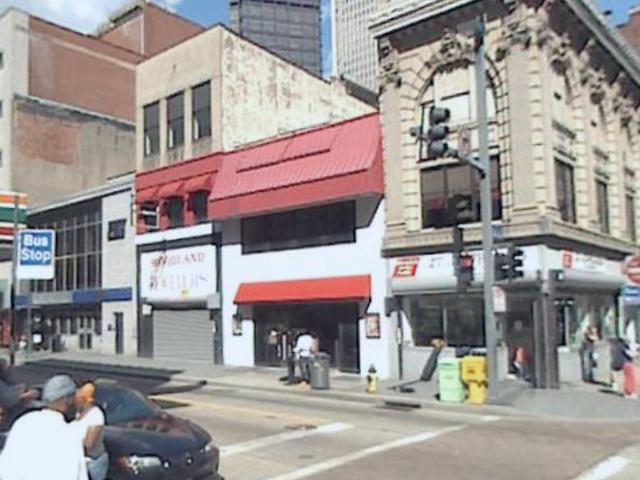} 
        \end{subfigure}
    \end{minipage}
    \begin{minipage}{.18\textwidth}
        \begin{subfigure}{\textwidth}
            \includegraphics[width=\textwidth]{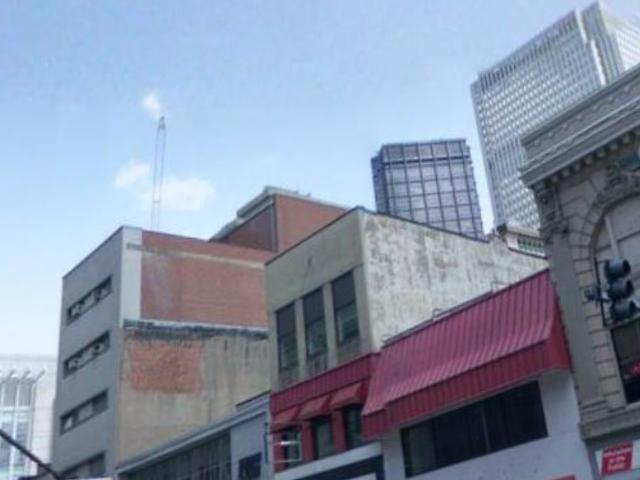}
        \end{subfigure}
        \begin{subfigure}{\textwidth}
            \includegraphics[width=\textwidth]{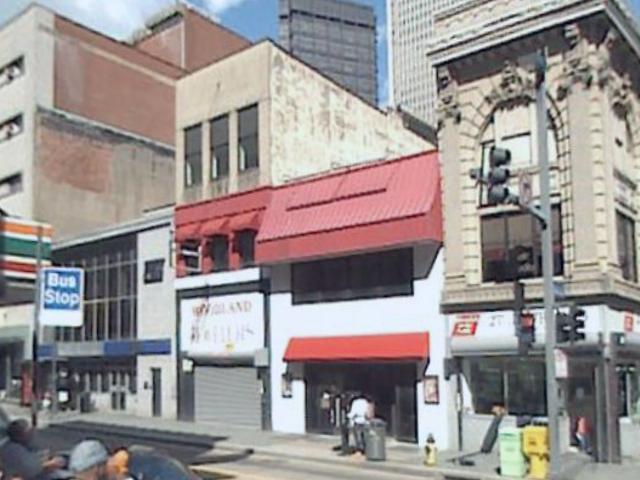} 
        \end{subfigure}
    \end{minipage}
    \begin{minipage}{.18\textwidth}
        \begin{subfigure}{\textwidth}
            \includegraphics[width=\textwidth]{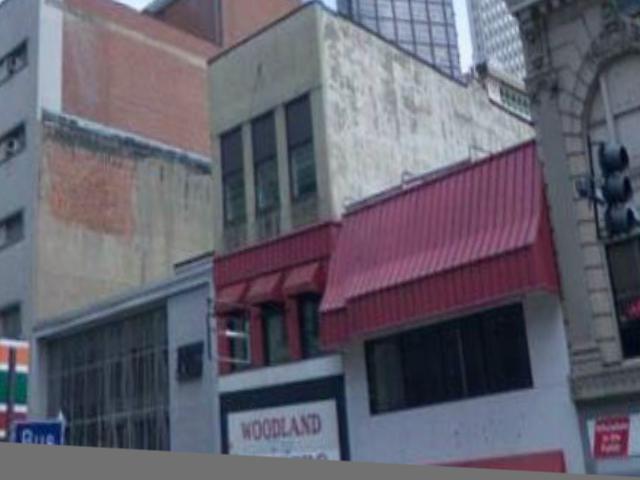}
        \end{subfigure}
        \begin{subfigure}{\textwidth}
            \includegraphics[width=\textwidth]{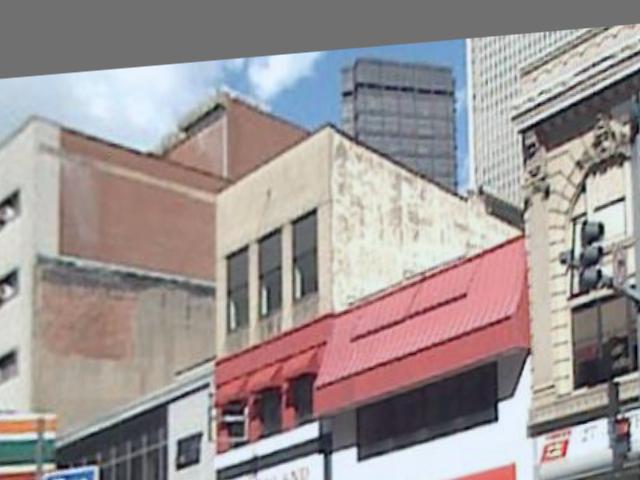} 
        \end{subfigure}
    \end{minipage}
    \resizebox{\textwidth}{!}{\includegraphics{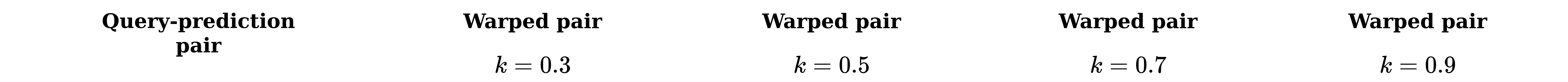}}
    \caption{Qualitative results: each pair of rows corresponds to one test case.
    The first column represents a query-prediction pair, the other columns show our pairwise warping's results using models trained with different values of $k$ (0.3, 0.5, 0.7, 0.9): a) the two images are 1 meter away, in this case a heavier warping is helpful; b) the two images are 39 meters away, and a heavy warping might be useful, depending on the chosen threshold for positive images; c) the prediction is wrongly retrieved by the global features, and warping has little to no effect on the warped pair, regardless of $k$'s value; d) a query-prediction pair with little visual overlap.}
    \label{fig:qualitative_k}
\end{figure*}

This effect on the generation of the training quadruplets $\{I_a, I_b, \boldsymbol{t}_a, \boldsymbol{t}_b\}$ translates to the fact that the network trained with higher values of $k$ learns to perform stronger warping.
We can see this qualitatively in Fig. \ref{fig:qualitative_k}, which showcases a few examples of the warped images generated by various query-prediction pairs, with our homography regression network trained using different values of $k$.
We can observe in Figs. \ref{fig:qualitative_k}a and \ref{fig:qualitative_k}d, that a network trained with small values of $k$ is not capable to compensate for strong viewpoint shifts and images with small overlap.
On the other hand, the model trained with high values of $k$ may become able to take two images of the same scene but at far away locations from each other and transform them to have similar appearance (see Fig. \ref{fig:qualitative_k}b). This explains why higher values of $k$ are better suited for the case of a rougher geolocalization.
Finally, Fig. \ref{fig:qualitative_k}c shows the result produced when the model is given a query and a false prediction. We can see that, regardless of the value of $k$, the warping operation has little effect. This demonstrates that our dense matching is rather robust to the case when the predictions to be re-ranked contain false positives.

\section{Comparison with other warping methods}
\begin{figure*}
    \centering
    \parbox[b]{.02\linewidth}{\subcaption{}}
    \begin{minipage}{.22\textwidth}
        \begin{subfigure}{\textwidth}
            \includegraphics[width=\textwidth]{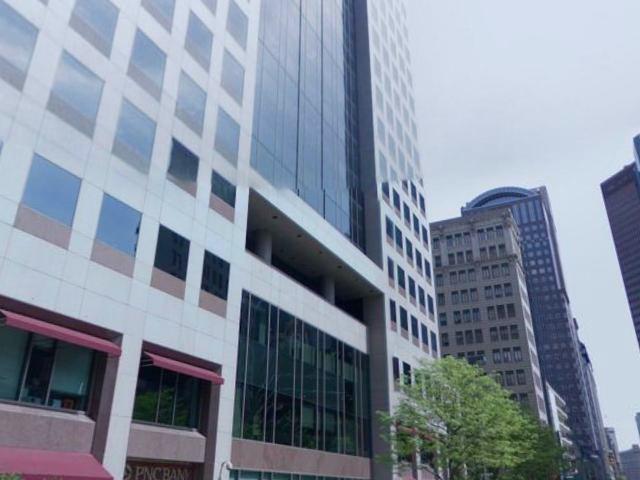}
        \end{subfigure}
        \begin{subfigure}{\textwidth}
            \includegraphics[width=\textwidth]{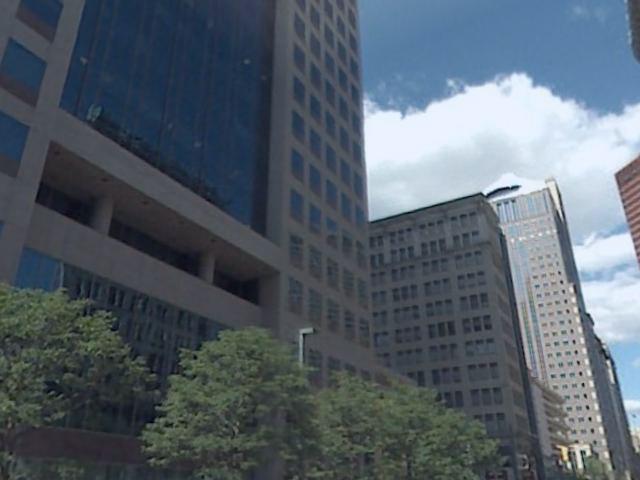} 
        \end{subfigure}
    \end{minipage}
    \hspace{2mm}\vline\hspace{2mm}
    \begin{minipage}{.22\textwidth}
        \begin{subfigure}{\textwidth}
            \includegraphics[width=\textwidth]{imgs/cmp_base0830_3_p.jpg}
        \end{subfigure}
        \begin{subfigure}{\textwidth}
            \includegraphics[width=\textwidth]{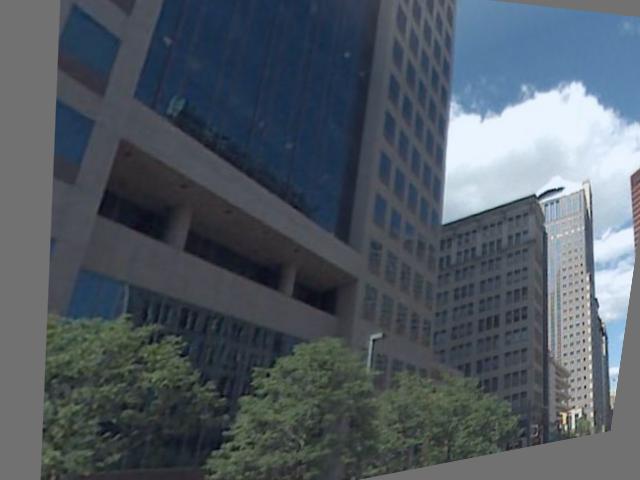} 
        \end{subfigure}
    \end{minipage}
    \begin{minipage}{.22\textwidth}
        \begin{subfigure}{\textwidth}
            \includegraphics[width=\textwidth]{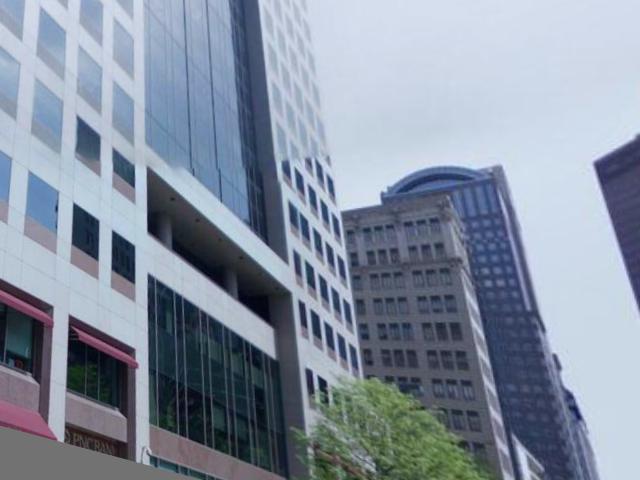}
        \end{subfigure}
        \begin{subfigure}{\textwidth}
            \includegraphics[width=\textwidth]{imgs/cmp_base0830_3_q.jpg} 
        \end{subfigure}
    \end{minipage}
    \begin{minipage}{.22\textwidth}
        \begin{subfigure}{\textwidth}
            \includegraphics[width=\textwidth]{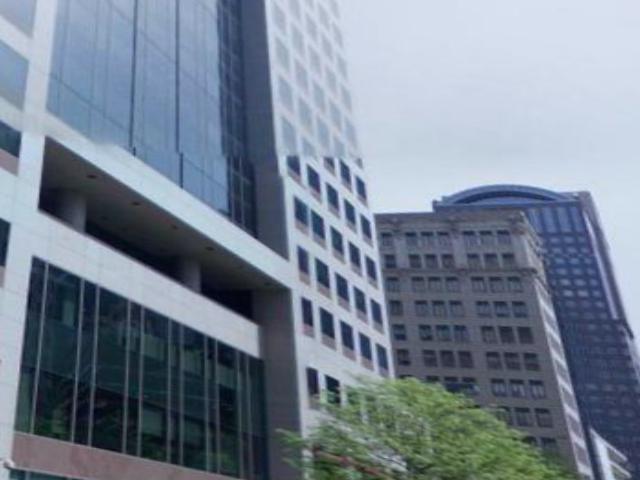}
        \end{subfigure}
        \begin{subfigure}{\textwidth}
            \includegraphics[width=\textwidth]{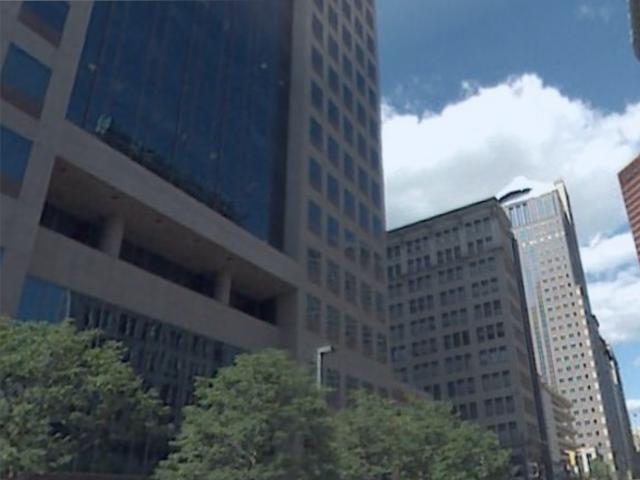} 
        \end{subfigure}
    \end{minipage}
    
    \vspace{3mm}
    \parbox[b]{.02\linewidth}{\subcaption{}}
    \begin{minipage}{.22\textwidth}
        \begin{subfigure}{\textwidth}
            \includegraphics[width=\textwidth]{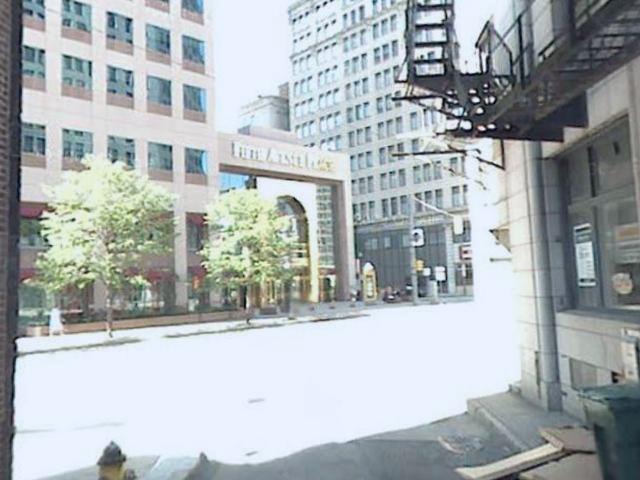}
        \end{subfigure}
        \begin{subfigure}{\textwidth}
            \includegraphics[width=\textwidth]{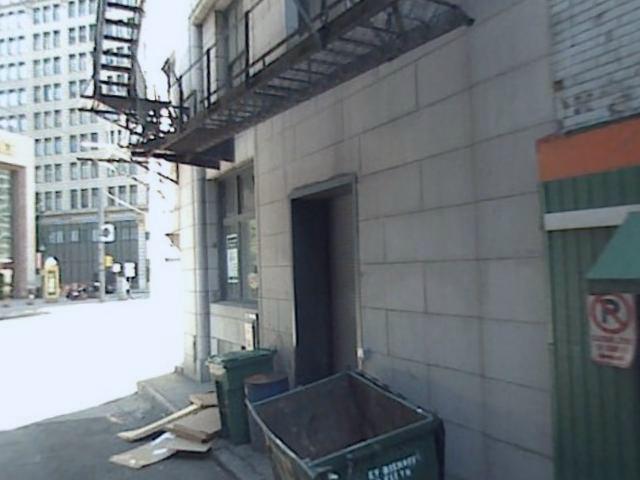} 
        \end{subfigure}
    \end{minipage}
    \hspace{2mm}\vline\hspace{2mm}
    \begin{minipage}{.22\textwidth}
        \begin{subfigure}{\textwidth}
            \includegraphics[width=\textwidth]{imgs/cmp_base1330_3_p.jpg}
        \end{subfigure}
        \begin{subfigure}{\textwidth}
            \includegraphics[width=\textwidth]{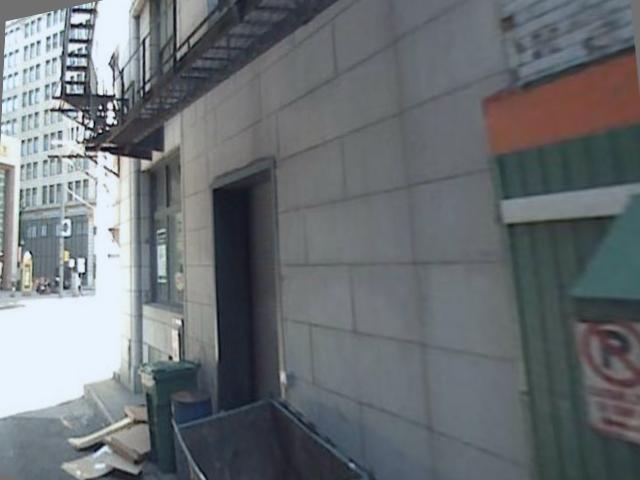} 
        \end{subfigure}
    \end{minipage}
    \begin{minipage}{.22\textwidth}
        \begin{subfigure}{\textwidth}
            \includegraphics[width=\textwidth]{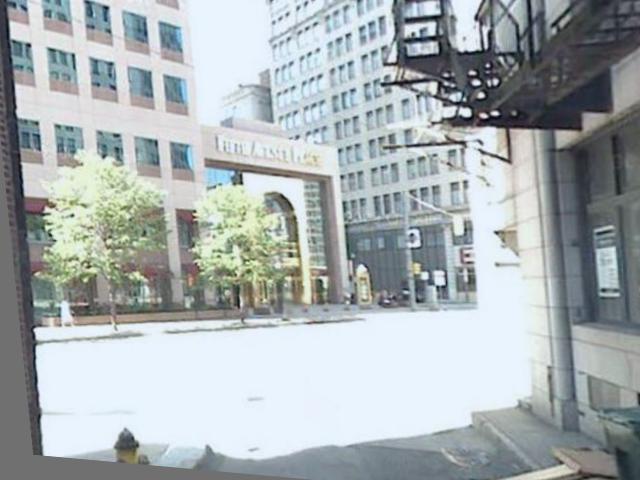}
        \end{subfigure}
        \begin{subfigure}{\textwidth}
            \includegraphics[width=\textwidth]{imgs/cmp_base1330_3_q.jpg} 
        \end{subfigure}
    \end{minipage}
    \begin{minipage}{.22\textwidth}
        \begin{subfigure}{\textwidth}
            \includegraphics[width=\textwidth]{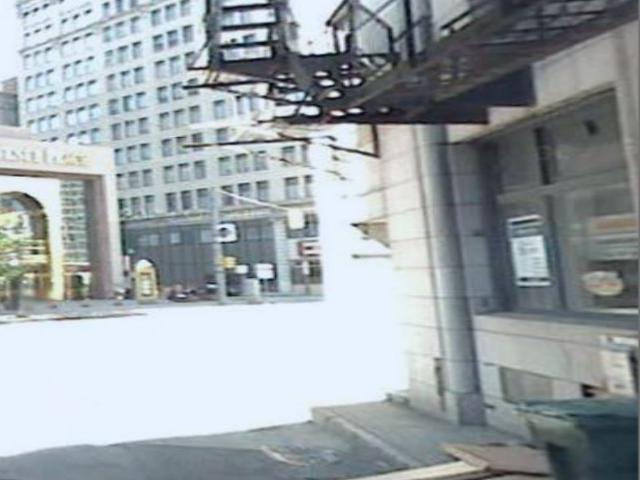}
        \end{subfigure}
        \begin{subfigure}{\textwidth}
            \includegraphics[width=\textwidth]{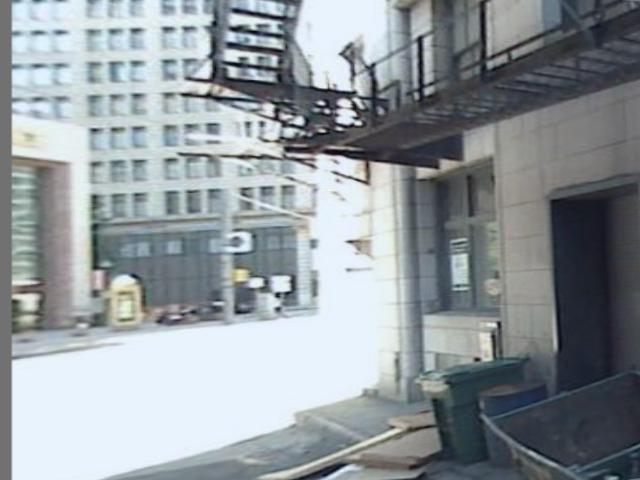} 
        \end{subfigure}
    \end{minipage}
    
    \vspace{3mm}
    \parbox[b]{.02\linewidth}{\subcaption{}}
    \begin{minipage}{.22\textwidth}
        \begin{subfigure}{\textwidth}
            \includegraphics[width=\textwidth]{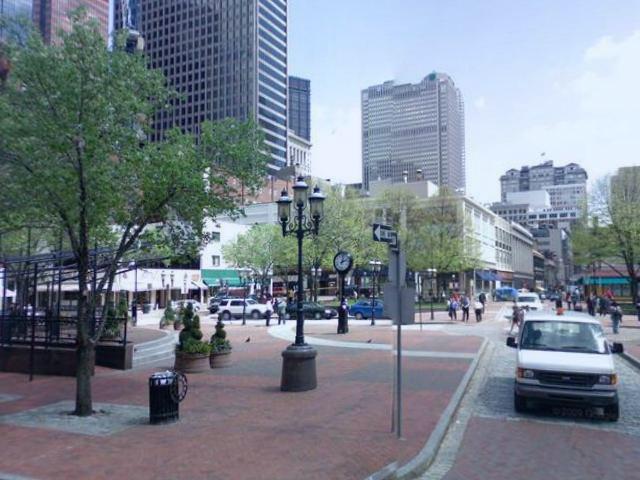}
        \end{subfigure}
        \begin{subfigure}{\textwidth}
            \includegraphics[width=\textwidth]{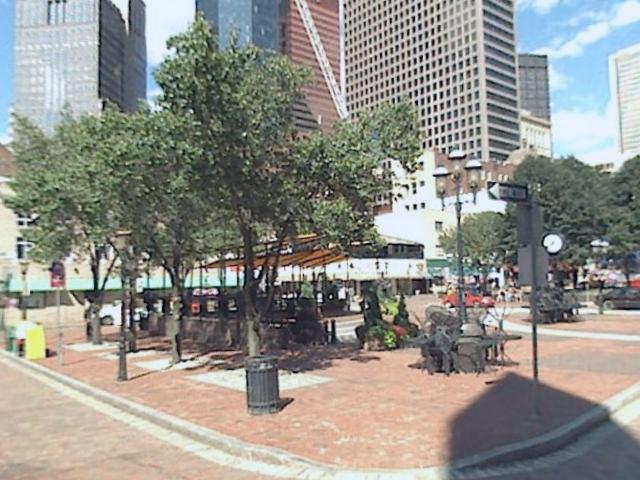} 
        \end{subfigure}
    \end{minipage}
    \hspace{2mm}\vline\hspace{2mm}
    \begin{minipage}{.22\textwidth}
        \begin{subfigure}{\textwidth}
            \includegraphics[width=\textwidth]{imgs/cmp_base1930_3_p.jpg}
        \end{subfigure}
        \begin{subfigure}{\textwidth}
            \includegraphics[width=\textwidth]{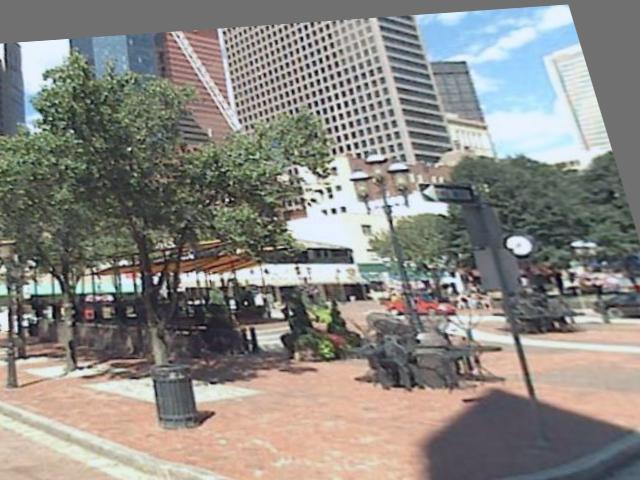} 
        \end{subfigure}
    \end{minipage}
    \begin{minipage}{.22\textwidth}
        \begin{subfigure}{\textwidth}
            \includegraphics[width=\textwidth]{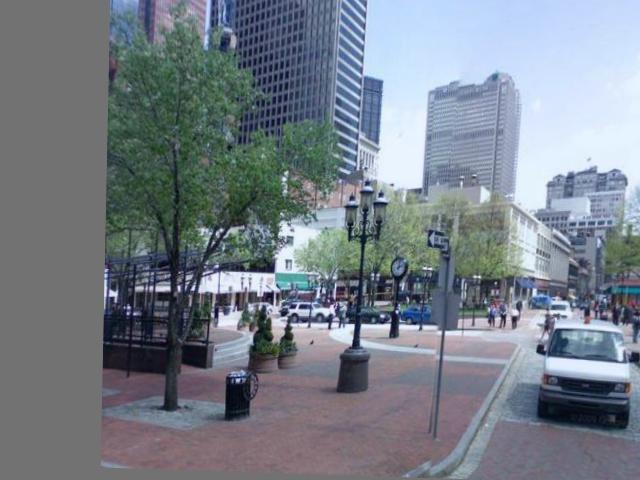}
        \end{subfigure}
        \begin{subfigure}{\textwidth}
            \includegraphics[width=\textwidth]{imgs/cmp_base1930_3_q.jpg} 
        \end{subfigure}
    \end{minipage}
    \begin{minipage}{.22\textwidth}
        \begin{subfigure}{\textwidth}
            \includegraphics[width=\textwidth]{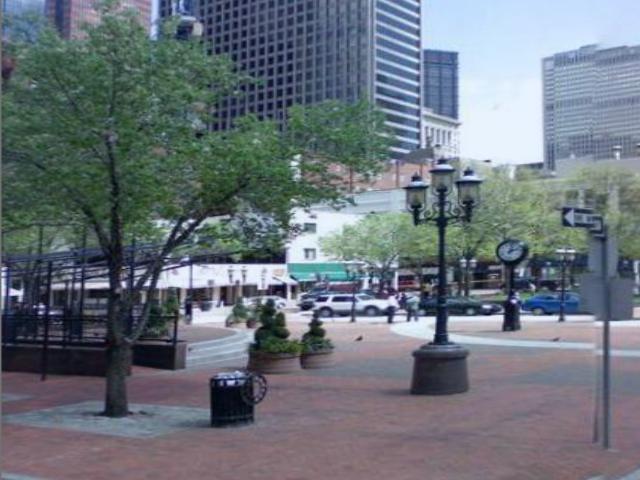}
        \end{subfigure}
        \begin{subfigure}{\textwidth}
            \includegraphics[width=\textwidth]{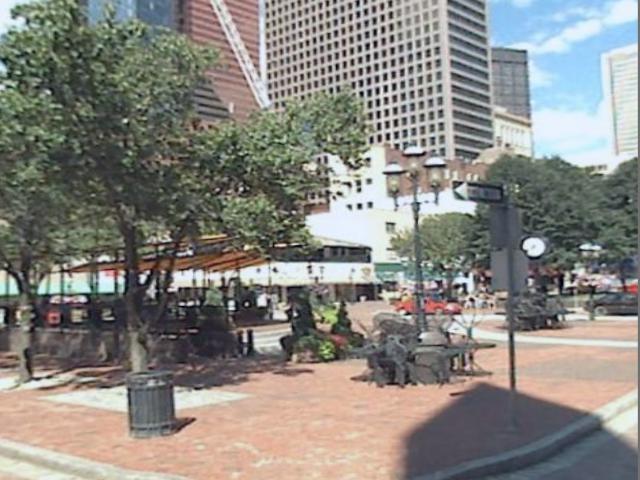} 
        \end{subfigure}
    \end{minipage}
    \resizebox{\textwidth}{!}{\includegraphics{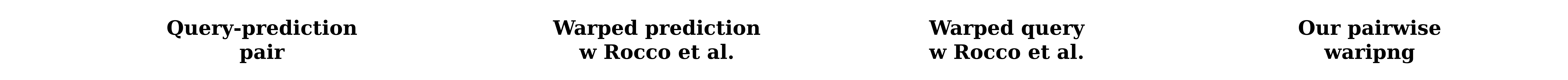}}
    \caption{Qualitative results of homography, each pair of rows corresponds to one test case. The first column represents a query-prediction pair, the second column shows warping on the prediction using \cite{rocco2018_warp}, the third shows warping on the query using \cite{rocco2018_warp}, and the rightmost column is the output of our pairwise warping, using our best network (ResNet-50 \cite{he2016_resnet} backbone trained with $k=0.6$.)}
    \label{fig:warping_comparisons}
\end{figure*}

In the main paper we have discussed how our warping regression module is inspired by the work of Rocco et al. \cite{rocco2018_warp}, detailing the conceptual differences and corroborating these considerations with quantitative and qualitative results. In Fig. \ref{fig:warping_comparisons} we provide further qualitative evidence to give a better intuition of the differences between the two methods.
Firstly, since \cite{rocco2018_warp} only transforms one image while keeping the other one unchanged, it gives rather different results depending on whether the transformed image is the query or the prediction (see Fig. \ref{fig:warping_comparisons}a). On the other hand, our method transforms both images achieving a higher robustness. 
Moreover, the single image transformation from \cite{rocco2018_warp} can lead to the creation of artifacts, as pixels outside of the source image boundaries are filled with grey color, which might further complicate the retrieval task (see Figs. \ref{fig:warping_comparisons}a and \ref{fig:warping_comparisons}c).
Finally, Fig. \ref{fig:warping_comparisons}b shows a difficult example with little overlap, in which GeoWarp clearly manages to output similar representations whereas \cite{rocco2018_warp} has almost no effect.

\end{document}